\documentclass[twocolumn]{article}

\usepackage{PRIMEarxiv}

\usepackage[utf8]{inputenc}
\usepackage[T1]{fontenc}
\usepackage{microtype}
\usepackage{graphicx}
\usepackage{booktabs}
\usepackage{multirow}
\usepackage{tabularx}
\usepackage{array}
\usepackage{arydshln}
\usepackage{amsmath,amssymb,amsfonts,mathtools}
\usepackage{algorithm}
\usepackage{algpseudocode}
\usepackage{tikz}
\usetikzlibrary{arrows.meta,positioning,calc,fit,shapes.geometric,backgrounds}
\usepackage{listings}
\usepackage{xcolor}
\lstdefinestyle{prompttext}{basicstyle=\ttfamily\footnotesize,breaklines=true,columns=fullflexible,keepspaces=true,showstringspaces=false}
\usepackage{natbib}
\usepackage{url}
\usepackage{hyperref}

\graphicspath{{figures/}}

\newenvironment{promptbox}{%
  \par\medskip\noindent\begin{minipage}{\linewidth}\hrule\medskip
}{%
  \medskip\hrule\end{minipage}\par\medskip
}
\newcommand{\promptdivider}{\par\smallskip\hrule\smallskip}
\newcommand{\promptheading}[1]{\textbf{#1}\par\smallskip}
\newcommand{\promptplaceholder}[1]{\texttt{\detokenize{#1}}}
\newcounter{listing}

\title{CUSP: Decomposable Collective Uncertainty for Multi-Agent Multimodal Reasoning}

\author{
Chung-En Johnny Yu\\
University of West Florida\\
Pensacola, Florida, USA\\
\texttt{cy31@students.uwf.edu}
\And
    David Garcia\\
    University of West Florida\\
    Pensacola, Florida, USA\\
    \texttt{djg41@students.uwf.edu} \\
\And
Brian Jalaian\\
University of West Florida\\
Pensacola, Florida, USA\\
\texttt{bjalaian@uwf.edu}
\And
Nathaniel D. Bastian\\
United States Military Academy\\
West Point, New York, USA\\
\texttt{nathaniel.bastian@westpoint.edu}
}

\begin{document}

\twocolumn[
\begin{@twocolumnfalse}
\maketitle
\begin{abstract}
Aggregating heterogeneous vision-language models (VLMs) can improve multimodal reasoning, but neither an individual model's confidence nor that of the aggregated answer measures reliability at the system level.
We present CUSP (Collective Uncertainty through Semantic Opinion Pooling), a training-free uncertainty quantification framework that maps multiple VLM responses to a shared semantic response space, pools them into a pooled semantic opinion, and reports two complementary system-level signals: collective uncertainty, the dispersion of the pooled opinion, and Jensen--Shannon divergence (JSD), the conflict among the model-level opinions.
Within this pooled semantic opinion, the unnormalized collective entropy decomposes exactly into the mean of the models' individual semantic entropies and the JSD, separating total dispersion from model conflict.
Requiring neither token logits nor calibration labels, CUSP applies to open-weight and commercial VLMs alike.
In static multi-VLM ensembles, collective uncertainty is the strongest signal in the small-model regime (0.764 AUROC for prediction-error detection, 0.889 AUARC for abstention), outperforming uncertainty baselines majority voting and naive selection by 4.7 to 15.8 points and widening its margin as the ensemble grows; JSD is strongest in the evaluated commercial regime (0.819 AUROC, 0.910 AUARC) and ranks hard-answer model conflict with AUROC up to 0.982.
The pooled prediction also improves accuracy over the average single model by 5.6 to 13.0 points.
Over the full trajectory of a multi-step, multi-agent system, subagent collective uncertainty ranks system failures above chance (0.619 AUROC) and gives the best abstention ordering among the evaluated signals (0.699 AUARC).
\end{abstract}
\vspace{0.5em}
\end{@twocolumnfalse}
]

\section{Introduction}
\label{sec:introduction}

Vision-language models (VLMs) are increasingly used in settings where errors carry real cost, such as medical decision support and multi-robot autonomy \citep{li2025survey_lvlm, chen2024unveiling}.
Uncertainty signals enable such systems to abstain, request verification, or gather more evidence before committing \citep{shorinwa2025survey, liu2025survey}.
Although combining heterogeneous VLMs can improve multimodal reasoning and robustness \citep{liu2025foundation, xie2024survey, chen2025ensemble}, aggregation can accumulate uncertainty: models may be individually uncertain or confidently conflict, and careless aggregation can amplify their errors \citep{lu2024merge, pan2025why}.
Neither single-model confidence nor confidence in the aggregated answer therefore characterizes system-level reliability.

Most UQ methods characterize one model and response, so per-model scores do not describe a combined system: closed-source APIs may restrict token probabilities \citep{finlayson2024logits}, and sampling-based estimates must be recomputed per model \citep{farquhar2024semantic, zhang2024vluncertainty}.
Ensemble methods aggregate parallel predictions and sometimes measure diversity, but do not separate total uncertainty into dispersion and conflict \citep{qu2025uqmerge, zhang2025consensus, muse2025}; agent and trajectory methods instead track the sequential steps of one agent \citep{han2024uala, saup2025}.
A system-level signal must align heterogeneous opinions in a shared semantic response space and distinguish pooled uncertainty from model conflict.

\begin{figure*}[t]
\centering
\includegraphics[width=1\textwidth]{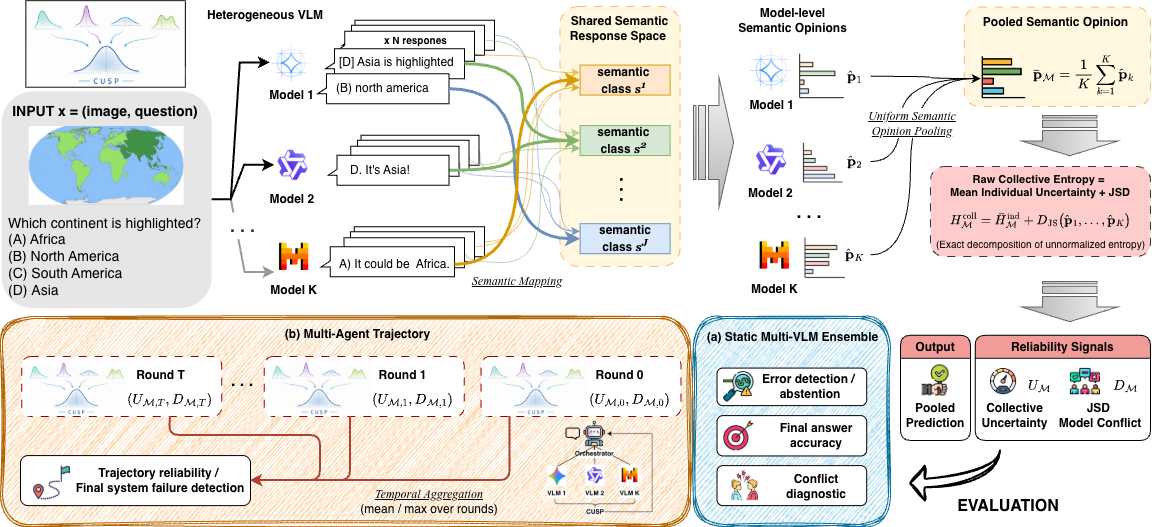}
\caption{Overview of CUSP. Sampled VLM responses are mapped into a shared semantic response space, pooled into a semantic opinion, and summarized by collective uncertainty and JSD.}
\label{fig:cusp-overview}
\end{figure*}

We introduce \textbf{CUSP (Collective Uncertainty through Semantic Opinion Pooling)}, a training-free framework for heterogeneous VLM systems (Figure~\ref{fig:cusp-overview}).
CUSP maps each model's sampled free-form responses to a shared semantic space, forms an empirical model-level opinion over meanings, and pools these opinions uniformly.
It reports collective uncertainty, the dispersion of the pooled opinion, and Jensen--Shannon divergence (JSD), conflict among model-level opinions, without requiring logits or calibration labels.

The unnormalized pooled entropy decomposes exactly into mean individual semantic entropy and JSD, separating average model uncertainty from conflict.
Systems can therefore have equal collective uncertainty because models are uniformly hesitant or confidently disagree, while JSD distinguishes these cases without serving as a proxy for prediction error.
We evaluate both signals in static ensembles and a multi-step, multi-agent evidence-acquisition trajectory, broadening validation beyond a single turn without claiming that CUSP controls the trajectory.

Our contributions are: (1) CUSP, a training-free semantic opinion-pooling framework producing collective uncertainty and JSD without logits or calibration labels from multiple heterogeneous VLMs; (2) an instantiation of the generalized JSD decomposition that separates pooled dispersion into mean individual uncertainty and model conflict; and (3) evaluation across static model regimes, ensemble sizes, and a multi-step multi-agent trajectory for prediction-error detection, abstention, and conflict diagnosis.

\section{Related Work}

\textbf{Uncertainty Quantification (UQ) for Individual LLMs and VLMs.}
UQ for individual LLMs includes token-probability and sampling-based approaches with different model-access requirements.
Token- and probability-based methods derive uncertainty from output token distributions and can flag unsupported claims \citep{fadeeva2024factchecking, ling2024decomposition}.
Commercial APIs may restrict token-probability information, while fuller logit access can expose proprietary model information \citep{finlayson2024logits}.
More fundamentally, over a long sequence, lexically different generations often express the same meaning, so token-probability dispersion does not track uncertainty at the semantic level \citep{farquhar2024semantic}.
Consistency-based methods estimate uncertainty from multiple sampled responses \citep{lin2023generating, xiong2023can}, and semantic entropy sharpens this idea by clustering semantic-similar generations before measuring dispersion \citep{farquhar2024semantic}.
These approaches suit black-box deployment because they need neither logits nor model training.

UQ for VLMs is comparatively underexplored and must contend with visual-evidence ambiguity in addition to language-generation variability \citep{li2024referencefree, xiao2025detecting, liu2025survey}.
Existing work often uses uncertainty to support training \citep{ji2023map}, decoding or refinement \citep{fang2024uncertainty, li2025mitigating}, or reasoning \citep{zhi2025seeing}, rather than reporting a model-level reliability signal.
The closest consistency-based method, VL-Uncertainty, detects hallucinations by sampling a VLM under perturbed multimodal inputs and adapts semantic entropy to the visual setting \citep{zhang2024vluncertainty}.
Across both modalities, these methods characterize a single model and a single response, and they leave open how uncertainty should be combined across several heterogeneous models.

\textbf{Multi-Model Ensemble Aggregation and Uncertainty.}
A separate line aggregates the outputs of several models to improve accuracy and robustness, through majority voting, confidence weighting, and larger heterogeneous ensembles \citep{li2024moreagents, ai2025beyond, lu2024merge, chen2025ensemble}.
Uncertainty at the ensemble level has received less attention, and the methods that address it stop short of a system-level reliability signal.
UQ-Merge uses uncertainty to guide multimodal aggregation without quantifying the uncertainty of the merged system \citep{qu2025uqmerge}, while Consensus Entropy weights VLM responses by uncertainty but evaluates neither system-level hallucination detection nor abstention \citep{zhang2025consensus}.
MUSE applies Jensen--Shannon divergence (JSD) to identify and aggregate a well-calibrated subset of LLMs \citep{muse2025}.
That use employs JSD as a selection-and-aggregation criterion; it does not relate the divergence to the total uncertainty of a pooled semantic opinion or decompose that uncertainty into mean individual uncertainty and conflict.
Reliability studies of multi-agent pipelines further show that naive aggregation can propagate individual errors into the shared output \citep{pan2025why}.
What remains missing is an aggregate uncertainty that separates the dispersion of the pooled semantic opinion from the conflict among the individual model opinions and admits an exact interpretation of the two.

\textbf{Trajectory-Level and Agent Uncertainty.}
Uncertainty in agentic systems shifts the question from a single response to how uncertainty accumulates across a multi-step trajectory.
Existing methods track this accumulation for one agent: UALA uses per-step uncertainty to govern when an agent acts or calls external tools \citep{han2024uala}, SAUP weights each step's uncertainty by situational importance as it propagates along the trajectory \citep{saup2025}, and information-theoretic and memory-based variants decompose or manage trajectory uncertainty in related ways \citep{uprop2025, auq2026}.
Recent multi-agent visual-reasoning systems use uncertainty-aware control, interaction monitoring, or agent diversity \citep{gamagent2025, guardian2025, multiagentblackbox2025}; a recent survey catalogs the open problems \citep{uqagents2026}.
These methods concentrate on the sequential states of a single agent or on final-answer reliability, and they do not jointly characterize the within-agent uncertainty of several concurrent heterogeneous agents, the conflict among their opinions at each round, and the risk carried across the trajectory.

We address this gap by constructing a shared semantic response space for heterogeneous VLMs and pooling their semantic opinions at the system level.
The unnormalized collective entropy decomposes exactly into mean individual uncertainty and JSD, giving the two signals a precise relationship.
We evaluate them in static multi-VLM ensembles and a multi-step, multi-agent evidence-acquisition trajectory.
\section{Method}
\label{sec:method}


We consider a system of $K$ off-the-shelf vision-language models (VLMs), $\mathcal{M}=\{M_1,\ldots,M_K\}$.
Each $M_k$ can operate as a standalone component in a static multi-model system or as an agent within a multi-agent system.
Given a multimodal input $x=(I,q)$, consisting of an image $I$ and a text query $q$, each model $M_k$ generates a fixed set of $N$ free-form responses, denoted by $\mathcal{Y}_k=\{y_k^1,\ldots,y_k^N\}$.
Each response is mapped into the shared semantic response space.
Semantic entropy was introduced to estimate uncertainty in free-form generations from a single LLM by grouping generations according to their semantic meanings \citep{kuhn2023semantic}.
We adapt this meaning-space construction to VLM outputs using empirical class frequencies.

\textbf{Mapping responses to a shared semantic response space.}
Let $\mathcal{Y}=\bigcup_{k=1}^K \mathcal{Y}_k$ denote the pooled response set.
A semantic mapping procedure constructs the shared semantic response space $\mathcal{S}_x=\{s^1,\ldots,s^J\}$.
In free-form tasks, an LLM-based semantic-equivalence judge clusters responses that express the same meaning.
Each $s^j$ denotes a discrete semantic-equivalence class of responses rather than a point in a continuous embedding space.
We represent the semantic clustering as a hard assignment function $g_x:\mathcal{Y}\rightarrow\{1,\ldots,J\}$, where $g_x(y)=j$ indicates that response $y$ is assigned to semantic class $s^j$.
Responses that differ in form but are semantically equivalent with respect to $x$ are mapped to the same class index.

\textbf{Forming each model's probabilistic opinion.}
For model $M_k$, we construct an empirical categorical distribution over the shared semantic response space.
The probability mass assigned to semantic response class $s^j$ is estimated by the relative frequency of the model's sampled responses assigned to that class:

\begin{equation}
\hat{p}_k(s^j\mid x)
=\frac{1}{N}\sum_{n=1}^{N}
\mathbf{1}\!\left[g_x(y_k^n)=j\right].
\label{eq:empirical-opinion}
\end{equation}

where $\mathbf{1}[\cdot]\in\{0,1\}$ is the indicator function.
This yields the probability vector
$\hat{\mathbf{p}}_k=
[\hat{p}_k(s^1\mid x),\ldots,\hat{p}_k(s^J\mid x)].$
For semantic classes absent from $M_k$'s sampled responses, the corresponding empirical probability mass is zero.
Because every sampled response is assigned to exactly one semantic class, all models are represented on the same support and each opinion is a valid categorical distribution:

\[
\hat p_k(s^j\mid x)\geq 0,
\qquad
\sum_{j=1}^{J}\hat p_k(s^j\mid x)=1.
\]

We quantify the uncertainty of model $M_k$ using the Shannon entropy of its empirical semantic opinion:

\begin{equation}
H_k
=-\sum_{j=1}^{J}
\hat p_k(s^j\mid x)
\log_2 \hat p_k(s^j\mid x),
\label{eq:individual-entropy}
\end{equation}

where $0\log_2 0:=0$.
A high $H_k$ indicates greater semantic variation among its sampled responses.
The mean individual uncertainty across the model set $\mathcal M$ is

\begin{equation}
\bar H_{\mathcal M}^{\mathrm{ind}}
=\frac{1}{K}\sum_{k=1}^{K}H_k.
\label{eq:mean-individual-entropy}
\end{equation}

\subsection{CUSP: Collective Uncertainty through Semantic Opinion Pooling}
\label{sec:cusp-pooling}

The representation above follows classical \emph{Linear Opinion Pooling (LOP)} \citep{stone1961opinion}, in which each model acts as a probabilistic expert that provides a probability distribution over a shared outcome space.
Applied to the generic expert distribution $\mathbf p_k$, LOP forms an aggregate distribution

\begin{equation}
\mathbf{p}_{\mathrm{LOP}}
=\sum_{k=1}^{K}w_k\mathbf{p}_k,
\qquad
w_k\ge 0,
\qquad
\sum_{k=1}^{K}w_k=1.
\label{eq:lop}
\end{equation}

Traditional LOP assumes that the experts already provide aligned probability vectors over a common support.
CUSP extends LOP to free-form outputs by constructing aligned semantic opinions for heterogeneous VLMs, which can then be pooled for response selection and uncertainty quantification.
CUSP uses uniform LOP weights, with $w_k=1/K$ for every model.
The pooled semantic opinion is

\begin{equation}
\bar{\mathbf p}_{\mathcal M}
=\frac{1}{K}\sum_{k=1}^{K}\hat{\mathbf{p}}_k
=\left[
\bar p_{\mathcal M}(s^1\mid x),
\ldots,
\bar p_{\mathcal M}(s^J\mid x)
\right],
\label{eq:pooled-opinion}
\end{equation}

where
$\bar p_{\mathcal M}(s^j\mid x)\geq 0$, 
$\sum_{j=1}^{J} \bar p_{\mathcal M}(s^j\mid x)=1.$
Uniform weighting preserves equal influence among the model-level opinions and does not assume that individual entropy estimates are calibrated or directly comparable across heterogeneous models.
Supplementary Section S6 evaluates an uncertainty-weighted pooling alternative.

The system selects a semantic response class having the largest probability under the collective opinion:

\begin{equation}
j^\star
\in
\arg\max_{j\in \{1,\ldots,J \}}
\bar p_{\mathcal M}(s^j\mid x),
\qquad
s^\star=s^{j^\star}.
\label{eq:pooled-selection}
\end{equation}

The system selects $y^\star=\textsc{Representative}(s^\star)$ and returns $y^\star$ as its pooled prediction.
We define the collective entropy of the model set $\mathcal M$ as the Shannon entropy of its pooled semantic opinion:

\begin{equation}
H_{\mathcal M}^{\mathrm{coll}}
=-\sum_{j=1}^{J}
\bar p_{\mathcal M}(s^j\mid x)
\log_2 \bar p_{\mathcal M}(s^j\mid x).
\label{eq:collective-entropy}
\end{equation}

Because a categorical distribution over $J$ semantic response classes has maximum entropy $\log_2J$, we report the normalized collective uncertainty

\begin{equation}
U_{\mathcal M}
=\begin{cases}
\dfrac{
H_{\mathcal M}^{\mathrm{coll}}
}{
\log_2 J
},
& J>1,
\\[4pt]
0,
& J=1.
\end{cases}
\label{eq:normalized-collective-uncertainty}
\end{equation}

Thus, $U_{\mathcal M}\in[0,1]$.
The normalization places collective uncertainty scores on a common scale across inputs that induce different numbers of semantic response classes.

\subsection{Decomposing Collective Uncertainty with JSD}
\label{sec:jsd-decomposition}

We begin from the standard generalized Jensen--Shannon divergence $D_{\mathrm{JS}}$, defined by Kullback--Leibler divergence $D_{\mathrm{KL}}$, for $K$ distributions $\mathbf p_1,\ldots,\mathbf p_K$ on a common discrete support and nonnegative weights $w_1,\ldots,w_K$ satisfying $\sum_k w_k=1$ \citep{lin1991divergence}.
Let $\mathbf p_w=\sum_k w_k\mathbf p_k$.
Then

\begin{equation}
\begin{aligned}
D_{\mathrm{JS},\mathbf w}(\mathbf p_1,\ldots,\mathbf p_K)
&=\sum_{k=1}^{K}w_kD_{\mathrm{KL}}\!\left(\mathbf p_k\,\middle\|\,\mathbf p_w\right)\\
&=H(\mathbf p_w)-\sum_{k=1}^{K}w_kH(\mathbf p_k).
\end{aligned}
\label{eq:generalized-jsd}
\end{equation}

CUSP uses $w_k=1/K$ and $\mathbf p_w=\bar{\mathbf p}_{\mathcal M}$.
Applying Equation~\eqref{eq:generalized-jsd} to the empirical semantic opinions gives

\begin{equation}
\begin{aligned}
D_{\mathrm{JS}}\!\left(\hat{\mathbf p}_1,\ldots,\hat{\mathbf p}_K\right)
&=\frac{1}{K}\sum_{k=1}^{K}D_{\mathrm{KL}}\!\left(\hat{\mathbf p}_k\,\middle\|\,\bar{\mathbf p}_{\mathcal M}\right)\\
&=H\!\left(\bar{\mathbf p}_{\mathcal M}\right)-\frac{1}{K}\sum_{k=1}^{K}H\!\left(\hat{\mathbf p}_k\right)\\
&=H_{\mathcal M}^{\mathrm{coll}}-\bar H_{\mathcal M}^{\mathrm{ind}}.
\end{aligned}
\label{eq:cusp-jsd}
\end{equation}

This yields the exact decomposition

\begin{equation}
H_{\mathcal M}^{\mathrm{coll}}
=
\bar H_{\mathcal M}^{\mathrm{ind}}
+
D_{\mathrm{JS}}\!\left(\hat{\mathbf p}_1,\ldots,\hat{\mathbf p}_K\right).
\label{eq:uncertainty-decomposition}
\end{equation}

It is well defined because all empirical opinions are categorical distributions on the shared finite support $\mathcal S_x$.
Moreover, if $\hat p_k(s^j\mid x)>0$, then $\bar p_{\mathcal M}(s^j\mid x)\geq \hat p_k(s^j\mid x)/K>0$, so no positive mass in a KL term is divided by zero.
For brevity, we write $D_{\mathrm{JS}}$ when its arguments are clear.
Because it is an average of KL divergences, $D_{\mathrm{JS}}\geq0$; it is zero if and only if all models assign identical probability mass to every semantic response class.
Within CUSP, JSD therefore quantifies model conflict through differences among semantic probability assignments.

For $K$ distributions over $J$ semantic response classes, the generalized JSD is upper-bounded by
$0 \leq D_{\mathrm{JS}} \leq \min \left( \log_2 K, \log_2 J \right)=\log_2 \min(K,J).$
We therefore define the normalized JSD score as

\begin{equation}
D_{\mathcal M}
:=
\begin{cases}
\dfrac{ D_{\mathrm{JS}} }{ \log_2 \min(K,J) }, & K>1 \text{ and } J>1,
\\[8pt]
0, & \text{otherwise}.
\end{cases}
\label{eq:normalized-jsd}
\end{equation}

Thus, $D_{\mathcal M}\in[0,1]$.
The denominator provides a common upper bound on the generalized JSD.
Equation~\eqref{eq:uncertainty-decomposition} is exact on the unnormalized entropy scale.
Collective uncertainty captures the total dispersion of the pooled semantic opinion, while JSD isolates the portion attributable to conflicts among the model opinions.
Low JSD does not imply correctness because all models may assign similar probability mass to the same incorrect response.

\begin{figure*}[t]
\centering
\begin{minipage}[t]{0.32\textwidth}
\centering
\includegraphics[width=\linewidth]{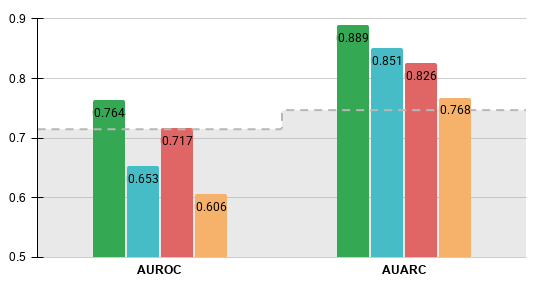}
\par{\footnotesize (a) Small open-source models}
\end{minipage}\hfill
\begin{minipage}[t]{0.32\textwidth}
\centering
\includegraphics[width=\linewidth]{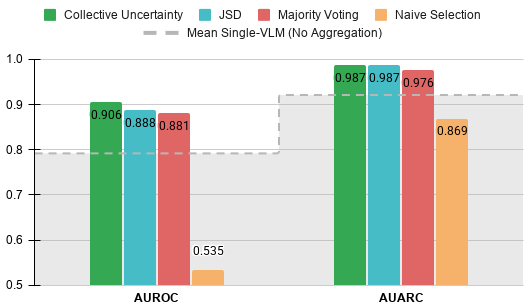}
\par{\footnotesize (b) Large open-source models}
\end{minipage}\hfill
\begin{minipage}[t]{0.32\textwidth}
\centering
\includegraphics[width=\linewidth]{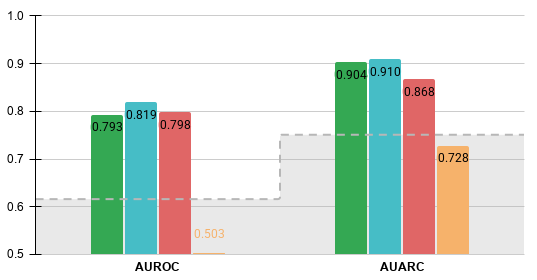}
\par{\footnotesize (c) Commercial models}
\end{minipage}
\caption{Static prediction-error detection (AUROC) and abstention (AUARC) across model regimes. Bars show combination-level means over 56 small, 20 large, and 20 commercial three-VLM systems; dashed segments show the mean single-VLM reference. Higher values are better.}
\label{fig:static-regimes}
\end{figure*}

The same definitions apply when the VLMs operate as subagents in a multi-agent system.
At interaction round $t$, the agents' responses induce a round-specific semantic response space $\mathcal S_{x_t}$ and empirical semantic opinions $\hat{\mathbf p}_{1,t}, \ldots, \hat{\mathbf p}_{K,t}$.
Applying Equations~\eqref{eq:pooled-opinion}--\eqref{eq:normalized-jsd} to these round-specific opinions yields the per-round collective uncertainty score $U_{\mathcal M,t}$ and normalized JSD score $D_{\mathcal M,t}$.
The multi-agent protocol and trajectory-level evaluation are specified in Section~\ref{sec:multi-agent-eval} and Supplementary Section S5.
Supplementary Algorithm S.1 in Supplementary Section S2 summarizes the complete CUSP procedure.

\section{Experiments}
\label{sec:experiments}

We evaluate the CUSP framework in two complementary settings.
For uncertainty quantification, we report its collective-uncertainty and JSD signals separately; for answer aggregation, we evaluate the pooled prediction.
Section~\ref{sec:ensemble-eval} treats multiple VLMs as a static ensemble and evaluates prediction-error detection, abstention ranking, and final-answer accuracy.
Section~\ref{sec:multi-agent-eval} places three heterogeneous VLM subagents for evidence extraction inside a centralized multi-step agentic reasoning system and asks whether the two evidence-layer signals can identify final system prediction failures.

\subsection{Multi-VLM Ensemble Evaluation}
\label{sec:ensemble-eval}

\textbf{Datasets.} 
We evaluate static multi-VLM aggregation on 400 samples from \emph{ScienceQA}, a multimodal scientific-reasoning benchmark \citep{lu2022scienceqa}, and 200 samples from \emph{MMMU-Pro}, a more robust expert-level multimodal reasoning benchmark \citep{yue2024mmmupro}.
\textbf{Metrics.} 
Unless otherwise specified, Area Under the ROC Curve (AUROC) measures how well an uncertainty score ranks incorrect above correct predictions, with 0.5 corresponding to random ranking. 
Area Under the Accuracy--Rejection Curve (AUARC) summarizes retained accuracy as increasingly uncertain predictions are rejected; higher values indicate a more useful uncertainty ranking for selective prediction. 
Section~\ref{sec:conflict-diagnostic} separately reports Conflict-AUROC, which uses hard-answer model conflict rather than prediction error as the positive-class label to quantify the ability of JSD to rank cases with model conflict.
Answer accuracy measures the quality of the aggregated prediction. 
Mean VLM UQ sampling latency at 50th-percentile (p50), denoted by $L_{\mathrm{sec}}$, is the summed time required for every VLM in a system to produce its uncertainty-estimation samples, averaged across systems of the same size. 
It is reported in seconds and increases as more VLMs are added. 
For each aggregation method, $\Delta L_{\mu\mathrm{s}}$ denotes the additional aggregation overhead above $L_{\mathrm{sec}}$ in microseconds; larger values indicate slower aggregation. 
Supplementary Section S4 gives the exact definitions and averaging procedures.
\textbf{Models and systems.}
We form three regimes: eight small open-source models (approximately 1--4B parameters), six large open-source models (approximately 27--35B parameters), and six commercial models.
Representative open-source models include Ministral3 \citep{mistral2025mistral3}, Gemma4 \citep{google2026gemma4}, and Qwen3.6 \citep{qwen2026qwen36}.
The commercial regime spans six major providers, including models such as Claude Sonnet 5 \citep{anthropic2026sonnet5}, Gemini 3.1 Pro Preview \citep{google2026gemini31}, and GPT-5.6 Terra \citep{openai2026gpt56}.
Supplementary Section S7 provides the complete model lists, versions, exact identifiers, and deployment details.
These regimes test whether CUSP remains useful across local model scales and ensemble sizes and with closed models whose decoding interfaces are less transparent.
Each VLM produces its main answer at prediction temperature 0.1, after which we sample $N=10$ additional responses at sampling temperature 1.0 to estimate the uncertainty. 
Supplementary Section S7 documents the limitation of some commercial APIs that do not accept the temperature parameter.
For a pool of $m$ models, prediction-error detection uses all $\binom{m}{3}$ three-VLM systems: 56 small, 20 large, and 20 commercial systems.
Accuracy and ensemble-size analyses use all subsets of size two or larger, totaling $\sum_{r=2}^{m}\binom{m}{r}$: 247 small, 57 large, and 57 commercial systems.
\textbf{Baselines and CUSP signals.} 
Naive Selection chooses the VLM with the lowest individual semantic entropy and uses that model's prediction and uncertainty. 
Majority Voting assigns one vote to each VLM's most probable semantic response and computes uncertainty from the normalized entropy of the empirical vote distribution. 
The average individual-VLM result is included as a non-aggregated reference rather than a system-level baseline. 
We evaluate CUSP's collective-uncertainty and JSD signals as components of the framework rather than as competing aggregation methods; Section~\ref{sec:jsd-decomposition} defines their distinct roles.
An uncertainty-weighted pooling variant is evaluated in Supplementary Section S6.

\textbf{Uncertainty Quality Across Model Regimes and Ensemble Sizes.}
\label{sec:static-uq}
Figure~\ref{fig:static-regimes} summarizes prediction-error detection across the three model regimes.
Collective uncertainty is strongest on both metrics in the small (0.764 AUROC/0.889 AUARC) and large (0.906/0.987) regimes, outperforming both baselines.
Higher AUROC shows that collective uncertainty ranks incorrect predictions more consistently, while higher AUARC shows that this ranking supports more effective abstention.
The commercial model regime presents a different pattern.
JSD is strongest at 0.819 AUROC and 0.910 AUARC, outperforming both baselines on both metrics.
Collective uncertainty trails Majority Voting by 0.5 AUROC points but exceeds it by 3.6 AUARC points.

As ensemble size varies (Supplementary Figure S.1), we examine how uncertainty quality changes in the small model regime as additional semantic opinions are pooled.
Majority Voting begins with a higher pairwise AUROC of 0.775 but falls below collective uncertainty from three models onward, while Naive Selection declines from 0.660 to 0.519 AUROC as the ensemble size grows.
Collective uncertainty increases from 0.757 to 0.840 AUROC over the same range, indicating that its prediction-error ranking benefits from additional heterogeneous opinions.
The JSD signal behaves differently: its prediction-error AUROC declines from 0.709 to 0.588 across the same sizes.
This is expected, because JSD diagnoses model conflict (Section~\ref{sec:conflict-diagnostic}) rather than prediction error, and the two signals therefore answer different questions.
\textbf{Pooling-policy ablation.}
We adopt uniform pooling because it admits the exact decomposition in Section~\ref{sec:jsd-decomposition} and improves as the ensemble grows; an uncertainty-weighted variant that assigns greater influence to models with lower individual uncertainty is competitive for small ensembles but scales less favorably, with complete static and trajectory results in Supplementary Section S6.

\textbf{Aggregated Accuracy and Latency.}
\label{sec:accuracy-latency}
Table~\ref{tab:accuracy} reports aggregated prediction accuracy.
CUSP is designed to aggregate semantic opinions and expose system reliability through collective uncertainty and JSD, rather than solely to optimize answer accuracy.
Nevertheless, the prediction selected from its pooled semantic opinion is more accurate than the main baselines in all three regimes.
Relative to the average individual VLM, the CUSP pooled prediction improves accuracy by 13.01, 5.56, and 8.70 percentage points in the small, large, and commercial regimes, respectively.
Across small-model systems of two to eight VLMs, CUSP's aggregation overhead remains at the microsecond scale and is negligible relative to the 2.04--8.30 seconds spent on VLM UQ sampling; the latency table appears in Supplementary Section S4.

\begin{table}[t]
\centering
\caption{Benchmark accuracy over all systems in each model regime; \textbf{bold} marks the best value per row.}
\label{tab:accuracy}
\footnotesize
\begin{tabular}{lr|rrr}
\toprule
Regime & \shortstack{Avg. single \\VLM} & CUSP & \shortstack{Majority\\voting} & \shortstack{Naive\\selection} \\
\midrule
Small      & 66.28\% & \textbf{79.29\%} & 71.86\% & 78.54\% \\
Large      & 88.42\% & \textbf{93.98\%} & 91.51\% & 89.14\% \\
Commercial & 71.00\% & \textbf{79.70\%} & 73.98\% & 71.60\% \\
\bottomrule
\end{tabular}
\end{table}

\textbf{JSD as a Model-Conflict Diagnostic.}
\label{sec:conflict-diagnostic}
Table~\ref{tab:conflict-auroc} reports conflict discrimination across model regimes.
Beyond prediction-error detection, we evaluate whether the two CUSP signals identify conflicts among model responses.
For each question, we assign a conflict label of zero when all models select the same top semantic response class and one when at least one model selects a different class.
We treat collective uncertainty and JSD as continuous ranking scores and report the resulting Conflict-AUROC.
JSD is the strongest conflict diagnostic in the small and large regimes and is within 0.005 of collective uncertainty in the commercial regime.
These results support JSD as a continuous diagnostic of hard-answer model conflict across the evaluated systems.
Because the positive class here is conflict rather than incorrect prediction, Conflict-AUROC is not interchangeable with the prediction-error detection AUROC reported in Section~\ref{sec:static-uq}.

\begin{table}[t]
\centering
\footnotesize
\caption{Hard-answer model-conflict discrimination across model regimes. Values report Conflict-AUROC; higher is better; \textbf{bold} marks the best value per row.}
\label{tab:conflict-auroc}
\begin{tabular}{lrr}
\toprule
Model regime & Collective uncertainty & JSD \\
\midrule
Small      & 0.872 & \textbf{0.916} \\
Large      & 0.962 & \textbf{0.982} \\
Commercial & \textbf{0.879} & 0.874 \\
\bottomrule
\end{tabular}
\end{table}

\subsection{Multi-Agent System Evaluation}
\label{sec:multi-agent-eval}

\textbf{Models and systems.}
The full system architecture, including the message-level procedure and prompts, is illustrated in Supplementary Figure S.2.
We evaluate a centralized multi-agent visual-reasoning system on 100 ScienceQA examples \citep{lu2022scienceqa}.
The system contains one text-only LFM2 (24B) orchestrator \citep{liquidai2026lfm224b} and three heterogeneous VLM subagents: Gemma4 (E2B) \citep{google2026gemma4}, Qwen3.5 (2B) \citep{qwen2026qwen35}, and Ministral3 (3B) \citep{mistral2025mistral3}.
A fixed Gemma4 captioner supplies general scene context.
The orchestrator performs three evidence-acquisition rounds: an initial general request (R0) followed by two task-targeted requests (R1 and R2).
At each round, it sends one focused instruction to the VLM subagents, each of which sees only the image and that instruction.
We intentionally use a compact centralized multi-agent system rather than a fully autonomous open-ended agent framework.
This design isolates a common structure in practical visual reasoning pipelines: a text-only orchestrator integrates evidence from multiple specialized VLM subagents across sequential acquisition rounds.
The simplified protocol keeps the decision path inspectable while preserving the core dependency we study: final system predictions are shaped by uncertain and potentially conflicting evidence from multiple visual agents.
Each VLM produces one main evidence response at temperature 0.1 and $N=10$ stochastic responses at temperature 0.9; only the ten stochastic responses form its semantic opinion.
The orchestrator and semantic-clustering calls use temperature 0.1.
LFM2 serves as the LLM-based semantic-clustering judge: it groups semantically equivalent free-form evidence responses into a shared response space.
It is an implementation component for semantic clustering, not an additional reasoning agent.
This setting deliberately measures uncertainty at the subagent layer rather than sampling the final orchestrator.
The orchestrator's prediction depends on the visual evidence supplied by these subagents; unreliable or ambiguous evidence, as well as evidence reflecting model conflict, can therefore propagate to the system output.
We test whether uncertainty in this evidence-collection layer ranks final system prediction failures, defined as an incorrect multiple-choice answer produced by the complete system.
\textbf{Baselines and CUSP signals.}
We evaluate CUSP's collective uncertainty ($U_{\mathcal{M}}$) and JSD ($D_{\mathcal{M}}$), defined in Section~\ref{sec:jsd-decomposition}, alongside four baselines.
MeanSE captures the mean within-model semantic entropy, MaxSE the most uncertain VLM, VoteEntropy the dispersion in the hard-vote distribution, and VoteConflict one minus the largest vote share.
In the static experiments, the Majority Voting baseline uses VoteEntropy as its uncertainty score.
These baselines test whether CUSP's signals add information beyond individual uncertainty summaries or hard decisions.
Exact definitions appear in Supplementary Section S5.
\textbf{Trajectory metrics.} Trajectory-level system-failure analysis compresses the three round-level scores using either their temporal mean or maximum. The mean captures uncertainty sustained across the trajectory, whereas the maximum captures its most uncertain round. Traj-AUROC evaluates ranking of incorrect versus correct final system predictions, and Traj-AUARC evaluates the corresponding abstention ranking. Round-AUROC instead pairs each round's uncertainty with a non-interventional same-round probe prediction formed from evidence available through that round; the probe does not alter the final trajectory. Formal definitions appear in Supplementary Sections S4 and S5, and prompts and implementation details in Supplementary Section S7.

\textbf{Trajectory-Level Failure Detection and Abstention.}
\label{sec:trajectory-results}
Table~\ref{tab:trajectory} compares temporal-mean and temporal-maximum trajectory summaries.
The final-round system answers 63 of the 100 examples correctly, leaving 37 system prediction failures for trajectory-level ranking.
Collective uncertainty provides the best abstention ordering under both temporal mean (0.694 Traj-AUARC) and maximum (0.699) aggregation.
VoteEntropy exceeds its mean Traj-AUROC by only 0.002 but has substantially lower mean Traj-AUARC (0.668 versus 0.694).
Under temporal maximum, collective uncertainty exceeds the strongest external baselines by 1.3 AUROC and 1.5 AUARC points, while JSD ranks second on AUARC under both temporal summaries.
Across the four mean/max comparisons, collective uncertainty is best in three; its Traj-AUROC remains moderate rather than near-perfect, but its consistent Traj-AUARC advantage supports prioritizing predictions for abstention or review.
In the round-wise probe analysis, collective uncertainty achieves the highest mean Round-AUROC (0.594), ahead of MeanSE (0.584) and JSD (0.580), although it does not lead at any individual round (Supplementary Section S5). Round-wise detectability declines as targeted evidence resolves visible ambiguity and residual errors shift toward evidence integration and reasoning.
Supplying the single CUSP-selected representative rather than the three separate evidence responses yields higher round-wise probe accuracy at every round ($+15.0$, $+1.0$, and $+3.0$ points at R0--R2); the per-round table appears in Supplementary Section S5.

\begin{table}[t]
\centering
\caption{Full-trajectory prediction-error detection (AUROC) and abstention ranking (AUARC) in the multi-agent system. Scores summarize the three evidence-acquisition rounds using temporal mean or maximum aggregation; \textbf{bold} and \underline{underline} mark the best and second-best values per column.}
\label{tab:trajectory}
\footnotesize
\begin{tabular}{ccccc}
\toprule
\multirow{2}{*}{Signal} & \multicolumn{2}{c}{Mean} & \multicolumn{2}{c}{Max} \\
\cmidrule(lr){2-3}\cmidrule(lr){4-5}
 & AUROC & AUARC & AUROC & AUARC \\
\midrule
\shortstack[c]{Collective\\uncertainty} & \underline{0.608} & \textbf{0.694} & \textbf{0.619} & \textbf{0.699} \\
JSD          & 0.603 & \underline{0.690} & 0.599 & \underline{0.689} \\
MeanSE       & 0.601 & 0.688 & 0.590 & 0.684 \\
MaxSE        & 0.530 & 0.668 & 0.489 & 0.652 \\
VoteEntropy  & \textbf{0.610} & 0.668 & \underline{0.606} & 0.666 \\
VoteConflict & 0.589 & 0.663 & 0.588 & 0.669 \\
\bottomrule
\end{tabular}
\end{table}
\section{Discussion}
\label{sec:discussion}

\textbf{Why Uniform Semantic Opinion Pooling?}
\label{sec:discussion-uniform-pooling}
The pooling-policy ablation in Section~\ref{sec:static-uq} shows that uncertainty weighting can improve some regimes, so we do not categorically reject confidence-dependent aggregation.
However, lower entropy indicates a concentrated opinion, not correctness; converting potentially miscalibrated confidence into aggregation influence assumes entropy is comparable across heterogeneous models and may give an overconfident but incorrect model disproportionate weight.
Uniform pooling avoids this calibration assumption, preserves equal influence among model-level opinions, and retains the exact decomposition of collective uncertainty into mean individual uncertainty and JSD.
The decomposition does not make uniform pooling optimal for every downstream task; rather, uniform pooling is a conservative default that separates collective reliability measurement from downstream policies that may deliberately assign different levels of trust.

\textbf{Practical Implications.}
\label{sec:practical-implications}
A low-collective-uncertainty, high-JSD case may pass an operational uncertainty threshold even when models favor different top semantic response classes.
Because the analysis does not establish this region's prevalence or prediction-error rate, we treat it as a diagnostic scenario rather than an observed failure category.
Trajectory scores support retrospective ranking of completed predictions for rejection or review, not online stopping, re-query, or abstention policies.

\textbf{Limitations.}
\label{sec:limitations.}
First, CUSP's collective uncertainty depends on semantic-clustering granularity and judge errors.
This limitation is more consequential in free-form multi-agent settings, which require an LLM-based judge, than in fixed-choice tasks with lightweight response parsing; the microsecond aggregation overhead reported for fixed-choice systems therefore does not include potentially non-negligible free-form semantic mapping.
Second, estimating each opinion from ten samples limits distributional resolution, and repeated VLM sampling dominates aggregation overhead while growing with the number of models and reasoning rounds; more sample-efficient black-box uncertainty estimators remain important future work.
Third, the commercial evaluation covers 20 three-VLM systems drawn from a selected pool of six models and should not be interpreted as a population estimate for commercial VLMs.
Finally, CUSP's signals do not cover every failure source: models may share a misconception, or the orchestrator may integrate consistent evidence incorrectly.
They are therefore complementary signals within a broader monitoring framework, not certificates of correctness.
\section{Conclusion}
\label{sec:conclusion}

We introduced CUSP, a training-free framework that pools sampled free-form VLM responses in a shared semantic space, with collective entropy decomposing exactly into mean individual uncertainty and JSD-based model conflict.
Across static ensembles, collective uncertainty supports prediction-error detection and abstention, JSD diagnoses conflict and is the strongest commercial-regime error signal, and in the multi-agent setting the subagent signals remain above chance with collective uncertainty providing the best trajectory-level abstention ordering.
Future work should reduce sampling and semantic-mapping costs and test whether the two signals can support online verification or evidence-acquisition policies.

\bibliographystyle{plainnat}
\bibliography{references}

@article{lu2022scienceqa,
  title={Learn to explain: Multimodal reasoning via thought chains for science question answering},
  author={Lu, Pan and Mishra, Swaroop and Xia, Tanglin and Qiu, Liang and Chang, Kai-Wei and Zhu, Song-Chun and Tafjord, Oyvind and Clark, Peter and Kalyan, Ashwin},
  journal={Advances in Neural Information Processing Systems},
  volume={35},
  pages={2507--2521},
  year={2022}
}

@inproceedings{li2025survey_lvlm,
  title={A Survey of State of the Art Large Vision Language Models: Benchmark Evaluations and Challenges},
  author={Li, Zongxia and Wu, Xiyang and Du, Hongyang and Liu, Fuxiao and Nghiem, Huy and Shi, Guangyao},
  booktitle={Proceedings of the Computer Vision and Pattern Recognition Conference},
  pages={1587--1606},
  year={2025}
}

@inproceedings{liu2025survey,
  title={Uncertainty quantification and confidence calibration in large language models: A survey},
  author={Liu, Xiaoou and Chen, Tiejin and Da, Longchao and Chen, Chacha and Lin, Zhen and Wei, Hua},
  booktitle={Proceedings of the 31st ACM SIGKDD Conference on Knowledge Discovery and Data Mining V. 2},
  pages={6107--6117},
  year={2025}
}

@article{fadeeva2024factchecking,
  title={Fact-checking the output of large language models via token-level uncertainty quantification},
  author={Fadeeva, Ekaterina and Rubashevskii, Aleksandr and Shelmanov, Artem and Petrakov, Sergey and Li, Haonan and Mubarak, Hamdy and Tsymbalov, Evgenii and Kuzmin, Gleb and Panchenko, Alexander and Baldwin, Timothy and others},
  journal={arXiv preprint arXiv:2403.04696},
  year={2024}
}

@article{ling2024decomposition,
  title={Uncertainty decomposition and quantification for in-context learning of large language models},
  author={Ling, Chen and Zhao, Xujiang and Cheng, Wei and Liu, Yanchi and Sun, Yiyou and Zhang, Xuchao and Oishi, Mika and Osaki, Takao and Matsuda, Katsushi and Ji, Jie and others},
  journal={CoRR},
  year={2024}
}

@article{farquhar2024semantic,
  title={Detecting hallucinations in large language models using semantic entropy},
  author={Farquhar, Sebastian and Kossen, Jannik and Kuhn, Lorenz and Gal, Yarin},
  journal={Nature},
  volume={630},
  number={8017},
  pages={625--630},
  year={2024},
  publisher={Nature Publishing Group UK London}
}

@article{zhang2024vluncertainty,
  title={Vl-uncertainty: Detecting hallucination in large vision-language model via uncertainty estimation},
  author={Zhang, Ruiyang and Zhang, Hu and Zheng, Zhedong},
  journal={arXiv preprint arXiv:2411.11919},
  year={2024}
}

@article{han2024uala,
  title={Towards uncertainty-aware language agent},
  author={Han, Jiuzhou and Buntine, Wray and Shareghi, Ehsan},
  journal={arXiv preprint arXiv:2401.14016},
  year={2024}
}

@article{liu2025foundation,
  title={Advances and challenges in foundation agents: From brain-inspired intelligence to evolutionary, collaborative, and safe systems},
  author={Liu, Bang and Li, Xinfeng and Zhang, Jiayi and Wang, Jinlin and He, Tanjin and Hong, Sirui and Liu, Hongzhang and Zhang, Shaokun and Song, Kaitao and Zhu, Kunlun and others},
  journal={arXiv preprint arXiv:2504.01990},
  year={2025}
}

@article{xie2024survey,
  title={Large multimodal agents: A survey},
  author={Xie, Junlin and Chen, Zhihong and Zhang, Ruifei and Wan, Xiang and Li, Guanbin},
  journal={arXiv preprint arXiv:2402.15116},
  year={2024}
}

@article{chen2025ensemble,
  title={Harnessing multiple large language models: A survey on llm ensemble},
  author={Chen, Zhijun and Li, Jingzheng and Chen, Pengpeng and Li, Zhuoran and Sun, Kai and Luo, Yuankai and Mao, Qianren and Yang, Dingqi and Sun, Hailong and Yu, Philip S},
  journal={arXiv preprint arXiv:2502.18036},
  year={2025}
}

@article{li2024moreagents,
  title={More agents is all you need},
  author={Li, Junyou and Zhang, Qin and Yu, Yangbin and Fu, Qiang and Ye, Deheng},
  journal={arXiv preprint arXiv:2402.05120},
  year={2024}
}

@article{li2024referencefree,
  title={Reference-free hallucination detection for large vision-language models},
  author={Li, Qing and Geng, Jiahui and Lyu, Chenyang and Zhu, Derui and Panov, Maxim and Karray, Fakhri},
  journal={arXiv preprint arXiv:2408.05767},
  year={2024}
}

@article{ai2025beyond,
  title={Beyond Majority Voting: LLM Aggregation by Leveraging Higher-Order Information},
  author={Ai, Rui and Pan, Yuqi and Simchi-Levi, David and Tambe, Milind and Xu, Haifeng},
  journal={arXiv preprint arXiv:2510.01499},
  year={2025}
}

@article{chen2024unveiling,
  title={Unveiling uncertainty: A deep dive into calibration and performance of multimodal large language models},
  author={Chen, Zijun and Hu, Wenbo and He, Guande and Deng, Zhijie and Zhang, Zheng and Hong, Richang},
  journal={arXiv preprint arXiv:2412.14660},
  year={2024}
}

@article{fang2024uncertainty,
  title={From uncertainty to trust: Enhancing reliability in vision-language models with uncertainty-guided dropout decoding},
  author={Fang, Yixiong and Yang, Ziran and Chen, Zhaorun and Zhao, Zhuokai and Zhou, Jiawei},
  journal={arXiv preprint arXiv:2412.06474},
  year={2024}
}

@article{finlayson2024logits,
  title={Logits of api-protected llms leak proprietary information},
  author={Finlayson, Matthew and Ren, Xiang and Swayamdipta, Swabha},
  journal={arXiv preprint arXiv:2403.09539},
  year={2024}
}

@inproceedings{ji2023map,
  title={Map: Multimodal uncertainty-aware vision-language pre-training model},
  author={Ji, Yatai and Wang, Junjie and Gong, Yuan and Zhang, Lin and Zhu, Yanru and Wang, Hongfa and Zhang, Jiaxing and Sakai, Tetsuya and Yang, Yujiu},
  booktitle={Proceedings of the IEEE/CVF conference on computer vision and pattern recognition},
  pages={23262--23271},
  year={2023}
}

@article{li2025mitigating,
  title={Mitigating Hallucinations in Large Vision-Language Models via Reasoning Uncertainty-guided Refinement},
  author={Li, Shenshen and Xu, Xing and Meng, Wenxin and Song, Jingkuan and Peng, Chong and Shen, Heng Tao},
  journal={IEEE Transactions on Multimedia},
  year={2025},
  publisher={IEEE}
}

@article{lin2023generating,
  title={Generating with confidence: Uncertainty quantification for black-box large language models},
  author={Lin, Zhen and Trivedi, Shubhendu and Sun, Jimeng},
  journal={arXiv preprint arXiv:2305.19187},
  year={2023}
}

@article{lu2024merge,
  title={Merge, ensemble, and cooperate! a survey on collaborative strategies in the era of large language models},
  author={Lu, Jinliang and Pang, Ziliang and Xiao, Min and Zhu, Yaochen and Xia, Rui and Zhang, Jiajun},
  journal={arXiv preprint arXiv:2407.06089},
  year={2024}
}

@article{shorinwa2025survey,
  title={A survey on uncertainty quantification of large language models: Taxonomy, open research challenges, and future directions},
  author={Shorinwa, Ola and Mei, Zhiting and Lidard, Justin and Ren, Allen Z and Majumdar, Anirudha},
  journal={ACM Computing Surveys},
  year={2025},
  publisher={ACM New York, NY}
}

@inproceedings{xiao2025detecting,
  title={Detecting and mitigating hallucination in large vision language models via fine-grained ai feedback},
  author={Xiao, Wenyi and Huang, Ziwei and Gan, Leilei and He, Wanggui and Li, Haoyuan and Yu, Zhelun and Shu, Fangxun and Jiang, Hao and Zhu, Linchao},
  booktitle={Proceedings of the AAAI Conference on Artificial Intelligence},
  volume={39},
  pages={25543--25551},
  year={2025}
}

@article{xiong2023can,
  title={Can llms express their uncertainty? an empirical evaluation of confidence elicitation in llms},
  author={Xiong, Miao and Hu, Zhiyuan and Lu, Xinyang and Li, Yifei and Fu, Jie and He, Junxian and Hooi, Bryan},
  journal={arXiv preprint arXiv:2306.13063},
  year={2023}
}

@article{zhang2025consensus,
  title={Consensus Entropy: Harnessing Multi-VLM Agreement for Self-Verifying and Self-Improving OCR},
  author={Zhang, Yulong and Liang, Tianyi and Huang, Xinyue and Cui, Erfei and Guo, Xu and Chu, Pei and Li, Chenhui and Zhang, Ru and Wang, Wenhai and Liu, Gongshen},
  journal={arXiv preprint arXiv:2504.11101},
  year={2025}
}

@article{zhi2025seeing,
  title={Seeing and Reasoning with Confidence: Supercharging Multimodal LLMs with an Uncertainty-Aware Agentic Framework},
  author={Zhi, Zhuo and Feng, Chen and Daneshmend, Adam and Orlu, Mine and Demosthenous, Andreas and Yin, Lu and Li, Da and Liu, Ziquan and Rodrigues, Miguel RD},
  journal={arXiv preprint arXiv:2503.08308},
  year={2025}
}

@inproceedings{kuhn2023semantic,
  title={Semantic Uncertainty: Linguistic Invariances for Uncertainty Estimation in Natural Language Generation},
  author={Kuhn, Lorenz and Gal, Yarin and Farquhar, Sebastian},
  booktitle={International Conference on Learning Representations},
  year={2023},
  url={https://openreview.net/forum?id=tWS-S_aRDRe}
}

@article{stone1961opinion,
  title={The Opinion Pool},
  author={Stone, Mervyn},
  journal={The Annals of Mathematical Statistics},
  volume={32},
  number={4},
  pages={1339--1342},
  year={1961},
  doi={10.1214/aoms/1177704873}
}

@article{lin1991divergence,
  title={Divergence Measures Based on the Shannon Entropy},
  author={Lin, Jianhua},
  journal={IEEE Transactions on Information Theory},
  volume={37},
  number={1},
  pages={145--151},
  year={1991},
  doi={10.1109/18.61115}
}

@article{hanley1982roc,
  title={The Meaning and Use of the Area under a Receiver Operating Characteristic ({ROC}) Curve},
  author={Hanley, James A. and McNeil, Barbara J.},
  journal={Radiology},
  volume={143},
  number={1},
  pages={29--36},
  year={1982},
  doi={10.1148/radiology.143.1.7063747}
}

@inproceedings{nadeem2009arc,
  title={Accuracy-Rejection Curves ({ARC}s) for Comparing Classification Methods with a Reject Option},
  author={Nadeem, Malik Sajjad Ahmed and Zucker, Jean-Daniel and Hanczar, Blaise},
  booktitle={Proceedings of the Third International Workshop on Machine Learning in Systems Biology},
  series={Proceedings of Machine Learning Research},
  volume={8},
  pages={65--81},
  year={2009},
  publisher={PMLR},
  url={https://proceedings.mlr.press/v8/nadeem10a.html}
}

@article{yue2024mmmupro,
  title={{MMMU-Pro}: A More Robust Multi-discipline Multimodal Understanding Benchmark},
  author={Yue, Xiang and Zheng, Tianyu and Ni, Yuansheng and Wang, Yubo and Zhang, Kai and Tong, Shengbang and Sun, Yuxuan and Yu, Botao and Zhang, Ge and Sun, Huan and Su, Yu and Chen, Wenhu and Neubig, Graham},
  journal={arXiv preprint arXiv:2409.02813},
  year={2024}
}

@inproceedings{qu2025uqmerge,
  title={{UQ-Merge}: Uncertainty Guided Multimodal Large Language Model Merging},
  author={Qu, Huaizhi and Zhao, Xinyu and Peng, Jie and Lee, Kwonjoon and Dariush, Behzad and Chen, Tianlong},
  booktitle={Findings of the Association for Computational Linguistics: ACL 2025},
  pages={1401--1417},
  year={2025},
  publisher={Association for Computational Linguistics},
  doi={10.18653/v1/2025.findings-acl.73},
  url={https://aclanthology.org/2025.findings-acl.73/}
}

@inproceedings{muse2025,
  title={Simple Yet Effective: An Information-Theoretic Approach to Multi-{LLM} Uncertainty Quantification},
  author={Kruse, Maya and Afshar, Majid and Khatwani, Saksham and Mayampurath, Anoop and Chen, Guanhua and Gao, Yanjun},
  booktitle={Proceedings of the 2025 Conference on Empirical Methods in Natural Language Processing},
  pages={30493--30504},
  year={2025},
  publisher={Association for Computational Linguistics},
  doi={10.18653/v1/2025.emnlp-main.1551},
  url={https://aclanthology.org/2025.emnlp-main.1551/}
}

@article{pan2025why,
  title={Why Do Multi-Agent {LLM} Systems Fail?},
  author={Cemri, Mert and Pan, Melissa Z. and Yang, Shuyi and Agrawal, Lakshya A. and Chopra, Bhavya and Tiwari, Rishabh and Keutzer, Kurt and Parameswaran, Aditya and Klein, Dan and Ramchandran, Kannan and Zaharia, Matei and Gonzalez, Joseph E. and Stoica, Ion},
  journal={arXiv preprint arXiv:2503.13657},
  year={2025}
}

@inproceedings{saup2025,
  title={Uncertainty Propagation on {LLM} Agent},
  author={Zhao, Qiwei and Li, Dong and Liu, Yanchi and Cheng, Wei and Sun, Yiyou and Oishi, Mika and Osaki, Takao and Matsuda, Katsushi and Yao, Huaxiu and Zhao, Chen and Chen, Haifeng and Zhao, Xujiang},
  booktitle={Proceedings of the 63rd Annual Meeting of the Association for Computational Linguistics (Volume 1: Long Papers)},
  pages={6064--6073},
  year={2025},
  publisher={Association for Computational Linguistics},
  doi={10.18653/v1/2025.acl-long.302},
  url={https://aclanthology.org/2025.acl-long.302/}
}

@article{uprop2025,
  title={{UProp}: Investigating the Uncertainty Propagation of {LLM}s in Multi-Step Agentic Decision-Making},
  author={Duan, Jinhao and Diffenderfer, James and Madireddy, Sandeep and Chen, Tianlong and Kailkhura, Bhavya and Xu, Kaidi},
  journal={arXiv preprint arXiv:2506.17419},
  year={2025}
}

@article{auq2026,
  title={Agentic Uncertainty Quantification},
  author={Zhang, Jiaxin and Choubey, Prafulla Kumar and Huang, Kung-Hsiang and Xiong, Caiming and Wu, Chien-Sheng},
  journal={arXiv preprint arXiv:2601.15703},
  year={2026}
}

@inproceedings{gamagent2025,
  title={{GAM-Agent}: Game-Theoretic and Uncertainty-Aware Collaboration for Complex Visual Reasoning},
  author={Zhang, Jusheng and Fan, Yijia and Lin, Wenjun and Chen, Ruiqi and Jiang, Haoyi and Chai, Wenhao and Wang, Jian and Wang, Keze},
  booktitle={Advances in Neural Information Processing Systems},
  volume={38},
  year={2025},
  url={https://proceedings.neurips.cc/paper_files/paper/2025/hash/d38512ba4121046784528ad9984af74a-Abstract-Conference.html}
}

@inproceedings{guardian2025,
  title={{GUARDIAN}: Safeguarding {LLM} Multi-Agent Collaborations with Temporal Graph Modeling},
  author={Zhou, Jialong and Wang, Lichao and Yang, Xiao},
  booktitle={Advances in Neural Information Processing Systems},
  volume={38},
  year={2025},
  url={https://proceedings.neurips.cc/paper_files/paper/2025/hash/0bc795afae289ed465a65a3b4b1f4eb7-Abstract-Conference.html}
}

@inproceedings{multiagentblackbox2025,
  title={Rethinking {LLM} Uncertainty: A Multi-Agent Approach to Estimating Black-Box Model Uncertainty},
  author={Feng, Yu and Htut, Phu Mon and Qi, Zheng and Xiao, Wei and Mager, Manuel and Pappas, Nikolaos and Halder, Kishaloy and Li, Yang and Benajiba, Yassine and Roth, Dan},
  booktitle={Findings of the Association for Computational Linguistics: EMNLP 2025},
  pages={12349--12375},
  year={2025},
  publisher={Association for Computational Linguistics},
  doi={10.18653/v1/2025.findings-emnlp.660},
  url={https://aclanthology.org/2025.findings-emnlp.660/}
}

@inproceedings{uqagents2026,
  title={Uncertainty Quantification in {LLM} Agents: Foundations, Emerging Challenges, and Opportunities},
  author={Oh, Changdae and Park, Seongheon and Kim, To Eun and Li, Jiatong and Li, Wendi and Yeh, Samuel and Du, Sean and Hassani, Hamed and Bogdan, Paul and Song, Dawn and Li, Sharon},
  booktitle={Proceedings of the 64th Annual Meeting of the Association for Computational Linguistics (Volume 1: Long Papers)},
  pages={16219--16250},
  year={2026},
  publisher={Association for Computational Linguistics},
  doi={10.18653/v1/2026.acl-long.738},
  url={https://aclanthology.org/2026.acl-long.738/}
}

@article{bai2025qwen25vl,
  title={{Qwen2.5-VL} Technical Report},
  author={Bai, Shuai and Chen, Keqin and Liu, Xuejing and Wang, Jialin and Ge, Wenbin and Song, Sibo and Dang, Kai and Wang, Peng and Wang, Shijie and Tang, Jun and Zhong, Humen and Zhu, Yuanzhi and Yang, Mingkun and Li, Zhaohai and Wan, Jianqiang and Wang, Pengfei and Ding, Wei and Fu, Zheren and Xu, Yiheng and Ye, Jiabo and Zhang, Xi and Xie, Tianbao and Cheng, Zesen and Zhang, Hang and Yang, Zhibo and Xu, Haiyang and Lin, Junyang},
  journal={arXiv preprint arXiv:2502.13923},
  year={2025}
}

@article{bai2025qwen3vl,
  title={{Qwen3-VL} Technical Report},
  author={Bai, Shuai and others},
  journal={arXiv preprint arXiv:2511.21631},
  year={2025}
}

@misc{mistral2025mistral3,
  author={{Mistral AI}},
  title={Mistral 3},
  year={2025},
  howpublished={\url{https://mistral.ai/news/mistral-3/}},
  note={Accessed: 2026-07-24}
}

@misc{ibm2025granitevision,
  author={{IBM Granite Team}},
  title={Granite Vision 3.2 2B Model Card},
  year={2025},
  howpublished={\url{https://huggingface.co/ibm-granite/granite-vision-3.2-2b}},
  note={Accessed: 2026-07-24}
}

@misc{qwen2026qwen35,
  author={{Qwen Team}},
  title={{Qwen3.5} Official Model Collection},
  year={2026},
  howpublished={\url{https://huggingface.co/collections/Qwen/qwen35}},
  note={Accessed: 2026-07-24}
}

@misc{qwen2026qwen36,
  author={{Qwen Team}},
  title={{Qwen3.6-35B-A3B} Model Card},
  year={2026},
  howpublished={\url{https://huggingface.co/Qwen/Qwen3.6-35B-A3B}},
  note={Accessed: 2026-07-24}
}

@misc{google2025gemma3,
  author={{Google DeepMind}},
  title={Gemma 3 Model Card},
  year={2025},
  howpublished={\url{https://ai.google.dev/gemma/docs/core/model_card_3}},
  note={Accessed: 2026-07-24}
}

@misc{google2026gemma4,
  author={{Google DeepMind}},
  title={Gemma 4 Model Documentation},
  year={2026},
  howpublished={\url{https://ai.google.dev/gemma/docs}},
  note={Accessed: 2026-07-24}
}

@misc{liu2024llavanext,
  author={Liu, Haotian and Li, Chunyuan and Li, Yuheng and Lee, Bo and Li, Yuanhan and Shen, Sheng and Lee, Yong Jae},
  title={{LLaVA-NeXT}: Improved Reasoning, {OCR}, and World Knowledge},
  year={2024},
  howpublished={\url{https://llava-vl.github.io/blog/2024-01-30-llava-next/}},
  note={Accessed: 2026-07-24}
}

@misc{ollama2024llavaphi3,
  author={{Ollama}},
  title={{LLaVA-Phi3} Model Page},
  year={2024},
  howpublished={\url{https://ollama.com/library/llava-phi3}},
  note={Accessed: 2026-07-24}
}

@misc{korrapati2024moondream2,
  author={Korrapati, Vik and others},
  title={Moondream2 Model Card},
  year={2024},
  howpublished={\url{https://huggingface.co/vikhyatk/moondream2}},
  note={Accessed: 2026-07-24}
}

@misc{nvidia2026nemotron3,
  author={{NVIDIA}},
  title={{NVIDIA Nemotron 3} Family of Models},
  year={2026},
  howpublished={\url{https://research.nvidia.com/labs/nemotron/Nemotron-3/}},
  note={Accessed: 2026-07-24}
}

@misc{ollama2026nemotron3,
  author={{Ollama}},
  title={{Nemotron 3} Deployment Tags},
  year={2026},
  howpublished={\url{https://ollama.com/library/nemotron3/tags}},
  note={Accessed: 2026-07-24}
}

@misc{liquidai2026lfm224b,
  author={{Liquid AI}},
  title={{LFM2-24B-A2B}: Scaling Up the {LFM2} Architecture},
  year={2026},
  howpublished={\url{https://www.liquid.ai/blog/lfm2-24b-a2b}},
  note={Accessed: 2026-07-24}
}

@misc{anthropic2026sonnet5,
  author={{Anthropic}},
  title={Claude Sonnet 5},
  year={2026},
  howpublished={\url{https://www.anthropic.com/news/claude-sonnet-5}},
  note={Accessed: 2026-07-24}
}

@misc{google2026gemini31,
  author={{Google}},
  title={Gemini 3.1 Pro Preview Model Documentation},
  year={2026},
  howpublished={\url{https://ai.google.dev/gemini-api/docs/models/gemini-3.1-pro-preview}},
  note={Accessed: 2026-07-24}
}

@misc{minimax2026m3,
  author={{MiniMax}},
  title={MiniMax M3},
  year={2026},
  howpublished={\url{https://www.minimax.io/models/text/m3}},
  note={Accessed: 2026-07-24}
}

@misc{moonshot2026kimi26,
  author={{Moonshot AI}},
  title={Kimi K2.6 Quickstart},
  year={2026},
  howpublished={\url{https://platform.kimi.ai/docs/guide/kimi-k2-6-quickstart}},
  note={Accessed: 2026-07-24}
}

@misc{openai2026gpt56,
  author={{OpenAI}},
  title={{GPT-5.6 Terra} Model Documentation},
  year={2026},
  howpublished={\url{https://developers.openai.com/api/docs/models/gpt-5.6-terra}},
  note={Accessed: 2026-07-24}
}

@misc{xai2026grok45,
  author={{xAI}},
  title={Grok 4.5 Model Documentation},
  year={2026},
  howpublished={\url{https://docs.x.ai/developers/models/grok-4.5}},
  note={Accessed: 2026-07-24}
}

\clearpage
\onecolumn
\setcounter{section}{0}
\renewcommand{\thesection}{S\arabic{section}}
\setcounter{figure}{0}
\renewcommand{\thefigure}{S.\arabic{figure}}
\setcounter{table}{0}
\renewcommand{\thetable}{S.\arabic{table}}
\setcounter{algorithm}{0}
\renewcommand{\thealgorithm}{S.\arabic{algorithm}}
\setcounter{listing}{0}
\renewcommand{\thelisting}{S.\arabic{listing}}

\section{Additional Notation}
\label{supp:notation}

For each evaluated question--image pair $x_i$, the index $i$ identifies one evaluation item.
We use $\mathbf p_{i,k}$ as item-indexed shorthand for model $M_k$'s empirical opinion $\hat{\mathbf p}_k$ evaluated on $x_i$.
Its support is the shared semantic response space $\mathcal S_{x_i}=\{s^1,\ldots,s^{J_i}\}$.
CUSP forms

\[
\bar{\mathbf p}_{\mathcal M,i}
=\frac{1}{K}\sum_{k=1}^{K}\mathbf p_{i,k},
\]

selects the top semantic response class, and reports the normalized collective uncertainty $U_{\mathcal M,i}$ and normalized JSD score $D_{\mathcal M,i}$.
These are the item-indexed counterparts of $\bar{\mathbf p}_{\mathcal M}$, $U_{\mathcal M}$, and $D_{\mathcal M}$ in Sections 3.1--3.2 of the main paper.
The adjective ``uniform'' is unnecessary in paper-facing signal names because equal model weights are part of the CUSP definition.

\section{CUSP Semantic Opinion Pooling Algorithm}
\label{supp:cusp-algorithm}

Algorithm~\ref{alg:cusp} summarizes the complete CUSP procedure described in Section 3 of the main paper.

\begin{algorithm}[t]
\caption{CUSP semantic opinion pooling}
\label{alg:cusp}
\footnotesize
\begin{algorithmic}[1]
\Require Input $x$; models $\mathcal M=\{M_k\}_{k=1}^{K}$;
sample count $N$
\Ensure $y^\star,\bar{\mathbf p}_{\mathcal M},
U_{\mathcal M},D_{\mathcal M}$

\For{$k=1,\ldots,K$}
    \State $\mathcal Y_k
    \gets \{y_k^n\sim M_k(\cdot\mid x)\}_{n=1}^{N}$
\EndFor

\State $(\mathcal S_x,g_x)
\gets \textsc{BuildSemanticSpace}
(\bigcup_{k=1}^{K}\mathcal Y_k,x)$
\State $J\gets|\mathcal S_x|$,
where $\mathcal S_x=\{s^1,\ldots,s^J\}$

\For{$k=1,\ldots,K$}
    \State $\hat{\mathbf p}_k[j]
    \gets N^{-1}\sum_{n=1}^{N}
    \mathbf 1[g_x(y_k^n)=j],
    \quad j\in[J]$
    \State $H_k\gets H(\hat{\mathbf p}_k)$
\EndFor

\State $\bar{\mathbf p}_{\mathcal M}
\gets K^{-1}\sum_{k=1}^{K}\hat{\mathbf p}_k$
\State $j^\star
\gets \arg\max_{j\in[J]}
\bar{\mathbf p}_{\mathcal M}[j]$
\State $y^\star
\gets \textsc{Representative}(s^{j^\star})$

\State $H_{\mathcal M}^{\mathrm{coll}}
\gets H(\bar{\mathbf p}_{\mathcal M})$
\State $U_{\mathcal M}
\gets H_{\mathcal M}^{\mathrm{coll}}/\log_2J$
\State $D_{\mathrm{JS}}
\gets H_{\mathcal M}^{\mathrm{coll}}
-K^{-1}\sum_{k=1}^{K}H_k$
\State $D_{\mathcal M}
\gets D_{\mathrm{JS}}/\log_2\min(K,J)$

\State \Return
$y^\star,\bar{\mathbf p}_{\mathcal M},
U_{\mathcal M},D_{\mathcal M}$
\end{algorithmic}
\end{algorithm}

\section{Baseline Definitions}
\label{supp:baselines}

\subsection{Majority Voting}
\label{supp:majority-voting}

Let each model's top semantic response class be

\[
z_{i,k}\in\arg\max_j p_{i,k}(s^j\mid x_i).
\]

The vote distribution and prediction are

\[
q_i(s^j)=\frac{1}{K}\sum_{k=1}^{K}\mathbf 1[z_{i,k}=j],
\qquad
\hat j_i^{\mathrm{MV}}\in\arg\max_j q_i(s^j).
\]

Its normalized uncertainty score is the entropy of the vote distribution,

\[
U_i^{\mathrm{vote}}=
\begin{cases}
-\dfrac{\sum_jq_i(s^j)\log_2q_i(s^j)}{\log_2J_i}, & J_i>1,\\[4pt]
0, & J_i=1.
\end{cases}
\]

In the static experiments, ties at either the per-model or aggregate stage are resolved by the first class in the benchmark answer-space order.
Thus, this deterministic implementation rule does not introduce an additional model call.

\subsection{Naive Selection}
\label{supp:naive-selection}

Naive Selection chooses the model with the lowest individual semantic entropy, where $H_{i,k}=H(\mathbf p_{i,k})$:

\[
k_i^*\in\arg\min_k H_{i,k},
\qquad
\hat j_i^{\mathrm{NS}}\in\arg\max_jp_{i,k_i^*}(s^j\mid x_i),
\]

and uses the selected model's normalized entropy as its uncertainty score:

\[
U_i^{\mathrm{NS}}=
\begin{cases}
\dfrac{H_{i,k_i^*}}{\log_2J_i}, & J_i>1,\\[4pt]
0, & J_i=1.
\end{cases}
\]

If multiple models share the minimum entropy, the static implementation selects the first model in the configured combination order; a class-probability tie is resolved by the first semantic response class.

\subsection{Average Individual VLM}
\label{supp:average-vlm}

The average-individual reference is the arithmetic mean of the component VLM accuracies or uncertainty scores, depending on the reported metric.
It does not produce an ensemble prediction and is therefore shown as a reference line rather than a competing aggregation rule.

\section{Evaluation Metrics}
\label{supp:evaluation-metrics}

Let $b_i\in\{0,1\}$ denote a task-specific positive-class label, and let $\rho_i$ be a continuous ranking score for which larger values indicate stronger evidence for the positive class.
Define $\mathcal P_b=\{i:b_i=1\}$ and $\mathcal N_b=\{i:b_i=0\}$.
When both sets are nonempty, AUROC can be written as the pairwise ranking probability

\[
\begin{aligned}
\mathrm{AUROC}(\rho;b)
&=\frac{1}{|\mathcal P_b||\mathcal N_b|}
\sum_{a\in\mathcal P_b}\sum_{d\in\mathcal N_b}\\
&\quad\left(\mathbf 1[\rho_a>\rho_d]+\frac{1}{2}\mathbf 1[\rho_a=\rho_d]\right).
\end{aligned}
\]

This pairwise expression is the standard probability-of-correct-ranking interpretation of AUROC \citep{hanley1982roc}.
An AUROC of 0.5 corresponds to chance ranking; higher values mean positive examples tend to receive larger scores.

\subsection{Prediction-error AUROC}
\label{supp:prediction-auroc}

Let $f_i=\mathbf 1[\hat y_i^{\mathrm{sys}}\neq y_i^*]$ indicate whether the system prediction is incorrect.
Unless otherwise specified, AUROC in the main paper refers to $\mathrm{AUROC}(\rho;f)$ and therefore measures prediction-error ranking.

\subsection{Model-Conflict AUROC}
\label{supp:conflict-auroc}

For question--system pair $(i,\mathcal M)$ with $|\mathcal M|\geq2$, let $z_{i,k}$ be model $k$'s top semantic response class, as in Supplementary Section~\ref{supp:baselines}.
We define hard-answer model conflict as

\[
c_{i,\mathcal M} =\mathbf 1\!\left[
\left|\left\{z_{i,k}:k\in\mathcal M\right\}\right|>1
\right].
\]

Thus, $c_{i,\mathcal M}=0$ only when all component models select the same semantic response class, and $c_{i,\mathcal M}=1$ when at least one top semantic response class differs.
For a ranking score $\rho_{i,\mathcal M}$, we report

\[
\operatorname{Conflict\text{-}AUROC}(\rho)
=\operatorname{AUROC}(\rho;c),
\]

where both $\rho$ and $c$ are indexed by the pooled question--system pairs $(i,\mathcal M)$.
Conflict-AUROC uses the standard AUROC calculation with a different binary target; it must not be interpreted as prediction-error detection.

\subsection{AUARC for abstention}
\label{supp:auarc}

Accuracy--rejection curves measure retained accuracy as increasingly uncertain predictions are rejected \citep{nadeem2009arc}.
For rejection rate $r$, remove the fraction $r$ of samples with the largest uncertainty scores and let $\mathcal R(r)$ denote the retained set.
Retained accuracy is

\[
\mathrm{Acc}(r)=\frac{1}{|\mathcal R(r)|}\sum_{i\in\mathcal R(r)}(1-f_i).
\]

Let $B$ be the number of evaluated items, and let $\pi$ sort these items from largest to smallest ranking score.
After rejecting the first $k$ items, the retained accuracy is

\[
A_k= 1-\frac{1}{B-k}\sum_{m=k+1}^{B}f_{\pi(m)},
\qquad k=0,\ldots,B-1.
\]

The static evaluation script uses the complete discrete rejection grid $r_k=k/B$ and applies the trapezoidal rule without extrapolating from $(B-1)/B$ to $1$:

\[
\mathrm{AUARC}
=\sum_{k=0}^{B-2}\frac{A_k+A_{k+1}}{2B}.
\]

Higher AUARC means that removing high-uncertainty predictions improves retained accuracy more rapidly.
It is an offline selective-reliability metric and does not by itself demonstrate an online abstention policy.

\subsection{Latency accounting}
\label{supp:latency}

Let $\ell_{i,k}$ be the measured inference latency for obtaining model $k$'s uncertainty-estimation samples on evaluation item $i$.
For each system, we first sum the component-model inference times and then take the 50th-percentile (p50) over evaluation items.
The reported VLM UQ sampling latency $L_{\mathrm{sec}}$ is the mean of these per-system values across all systems containing the same number of VLMs:

\[
L_{\mathrm{sec}} =\operatorname{mean}_{\mathrm{systems}}
\left[
\operatorname{median}_{i}
\left(\sum_k\ell_{i,k}\right)
\right].
\]

Thus, $L_{\mathrm{sec}}$ contains the uncertainty-estimation inference time of every VLM but no aggregation computation.
It grows with system size because each added VLM contributes another set of uncertainty samples.

For each aggregation method, $\Delta L_{\mu\mathrm{s}}$ is its mean additional aggregation overhead above the VLM UQ sampling latency.
If end-to-end latency is expressed in seconds, the corresponding total is

\[
L_{\mathrm{total}} =L_{\mathrm{sec}}+10^{-6}\Delta L_{\mu\mathrm{s}}.
\]

Accordingly, every aggregation overhead in the table is positive, and a larger $\Delta L_{\mu\mathrm{s}}$ indicates slower aggregation.
The released artifact stores the uncertainty-weighted pooling latency in seconds and the other methods as microsecond differences relative to that value.
We recover each absolute overhead above $L_{\mathrm{sec}}$ from those stored measurements; no new timing run is introduced.
Because aggregation latency is separately micro-benchmarked and may vary slightly across fresh runs, sub-microsecond ordering differences should not be interpreted as practically meaningful speed differences.

\begin{table}[t]
\centering
\footnotesize
\caption{Mean p50 latency for small-model systems. $L_{\mathrm{sec}}$ is VLM UQ sampling latency in seconds per sample, and $\Delta L_{\mu\mathrm{s}}$ is additional aggregation overhead in microseconds; lower is faster; \textbf{bold} marks the best value per row.}
\label{tab:latency}
\begin{tabular}{cc|ccc}
\toprule
\multirow{2}{*}{$K$-VLM} & \multirow{2}{*}{$L_{\mathrm{sec}}\downarrow$} & \multicolumn{3}{c}{\shortstack{Additional aggregation\\overhead $\Delta L_{\mu\mathrm{s}}\downarrow$}} \\
\cmidrule(lr){3-5}
 & & CUSP & \shortstack{Majority\\Voting} & \shortstack{Naive\\Selection} \\
\midrule
2 & 2.04 & 2.56 & 1.94 & \textbf{1.84} \\
3 & 3.07 & 3.06 & \textbf{2.22} & 2.48 \\
4 & 4.11 & 3.56 & \textbf{2.50} & 3.11 \\
5 & 5.15 & 4.03 & \textbf{2.74} & 3.72 \\
6 & 6.19 & 4.49 & \textbf{2.97} & 4.34 \\
7 & 7.24 & 4.95 & \textbf{3.21} & 4.89 \\
8 & 8.30 & 5.11 & \textbf{3.32} & 5.30 \\
\bottomrule
\end{tabular}
\end{table}

The VLM UQ sampling latency $L_{\mathrm{sec}}$ grows approximately additively because every additional VLM must generate its own uncertainty-estimation samples.
It increases from 2.04 seconds for two VLMs to 8.30 seconds for eight VLMs.
CUSP incurs the largest mean aggregation overhead across the three reported aggregation methods, averaging 3.97 microseconds, compared with 2.70 microseconds for Majority Voting and 3.67 microseconds for Naive Selection.
Even CUSP's largest measured overhead is only 5.11 microseconds, which is negligible relative to the 2.04--8.30 seconds spent obtaining the individual VLM uncertainty estimates.
CUSP therefore provides the higher-quality uncertainty estimates reported in Section 4.1.1 of the main paper without materially increasing end-to-end system latency.

\section{Multi-Agent Trajectory Metrics}
\label{supp:trajectory-metrics}

At round $t$, let $\mathbf p_{i,k,t}$ denote VLM $k$'s empirical opinion over the round-specific shared semantic response space containing $J_{i,t}$ classes.
Its normalized semantic entropy is

\[
H_{i,k,t}^{\mathrm{norm}}=
\begin{cases}
\dfrac{H(\mathbf p_{i,k,t})}{\log_2J_{i,t}}, & J_{i,t}>1,\\[4pt]
0, & J_{i,t}=1.
\end{cases}
\]

The individual-uncertainty baselines are

\[
\begin{aligned}
\mathrm{MeanSE}_{i,t}&=\frac{1}{K}\sum_{k=1}^{K}H_{i,k,t}^{\mathrm{norm}},\\
\mathrm{MaxSE}_{i,t}&=\max_{1\le k\le K}H_{i,k,t}^{\mathrm{norm}}.
\end{aligned}
\]

Let $z_{i,k,t}$ be VLM $k$'s top semantic response class and let $q_{i,t}(s^j)=K^{-1}\sum_k\mathbf 1[z_{i,k,t}=j]$ be the hard-vote distribution.
VoteEntropy is the normalized entropy of $q_{i,t}$, as in Supplementary Section~\ref{supp:baselines}.
VoteConflict is

\[
V_{i,t}^{\mathrm{conflict}}=1-\max_jq_{i,t}(s^j).
\]

MeanSE and MaxSE test whether within-model semantic uncertainty alone explains system failures; VoteEntropy and VoteConflict test whether hard decisions suffice without CUSP's probability-level collective-uncertainty and JSD signals.

For evaluation item $i$, let $\rho_{i,t}$ be an uncertainty ranking score observed at round $t\in\{0,\ldots,T_i-1\}$, matching the R0--R2 indexing in Section 4.2 of the main paper.
We summarize the completed trajectory using

\[
\rho_i^{\mathrm{mean}}=\frac{1}{T_i}\sum_{t=0}^{T_i-1}\rho_{i,t},
\qquad
\rho_i^{\max}=\max_{0\le t<T_i}\rho_{i,t}.
\]

Traj-AUROC applies the definition in Supplementary Section~\ref{supp:evaluation-metrics} to $\rho_i^{\mathrm{mean}}$ or $\rho_i^{\max}$ using the final orchestrator failure label.
Traj-AUARC sorts and rejects completed trajectories using the same scalar scores.
Round-AUROC instead pairs $\rho_{i,t}$ with the correctness of the non-interventional probe answer produced from evidence available through round $t$.

\subsection{Round-wise failure detection results}
\label{supp:round-wise-results}

\begin{table}[t]
\centering
\caption{Round-wise prediction-error detection (Round-AUROC) for the non-interventional probe answer ($n=100$). The average column is computed from full-precision R0--R2 values; bold and underline mark the best and second-best values per column.}
\label{tab:round-auroc}
\footnotesize
\begin{tabular}{lrrrr}
\toprule
Signal & R0 & R1 & R2 & Avg. \\
\midrule
Collective uncertainty & \underline{0.634} & \underline{0.598} & \underline{0.551} & \textbf{0.594} \\
JSD          & 0.626 & 0.576 & 0.536 & 0.580 \\
MeanSE       & 0.598 & \textbf{0.601} & \textbf{0.552} & \underline{0.584} \\
MaxSE        & \textbf{0.647} & 0.592 & 0.480 & 0.573 \\
VoteEntropy  & 0.610 & 0.529 & 0.518 & 0.552 \\
VoteConflict & 0.608 & 0.537 & 0.524 & 0.557 \\
\bottomrule
\end{tabular}
\end{table}

Table~\ref{tab:round-auroc} reports the per-round and average probe-error rankings.
The strongest single-round signal changes over the trajectory.
MaxSE leads at R0 with 0.647 Round-AUROC, while MeanSE leads at R1 and R2 with 0.601 and 0.552.
Collective uncertainty is second at every round, obtaining 0.634, 0.598, and 0.551.
Its mean Round-AUROC of 0.594 is consequently the highest among the six signals, ahead of MeanSE at 0.584 and JSD at 0.580.
Collective uncertainty is therefore the most consistently competitive estimator in this comparison, even though it is not the strongest signal at any individual round.
This consistency refers to its relative ranking across the three rounds, not to lower numerical variance or statistical significance.

Absolute Round-AUROC generally decreases as the system acquires targeted evidence.
Targeted acquisition may resolve visually observable ambiguity, leaving final errors driven by evidence integration, reasoning, or errors shared across subagents.
Under such conditions, the remaining failures need not produce high JSD or entropy in the evidence responses.

\subsection{Round-wise Probe Accuracy with CUSP-Aggregated Evidence}
\label{supp:probe-accuracy}

\begin{table}[t]
\centering
\footnotesize
\caption{Round-wise probe-answer accuracy with CUSP-aggregated and non-aggregated evidence. Accuracy is computed over all 100 examples per round, and gains are percentage points.}
\label{tab:aggregated-evidence}
\begin{tabular}{lrrr}
\toprule
Round & CUSP-aggregated & Non-aggregated & Gain \\
\midrule
R0 & 64.0\% & 49.0\% & $+15.0$ \\
R1 & 61.0\% & 60.0\% & $+1.0$ \\
R2 & 63.0\% & 60.0\% & $+3.0$ \\
\bottomrule
\end{tabular}
\end{table}

Table~\ref{tab:aggregated-evidence} compares probe accuracy with CUSP-aggregated and non-aggregated evidence.
The aggregated condition supplies the orchestrator with one representative selected from CUSP's pooled distribution; the matched non-aggregated condition supplies the three main evidence responses separately.
Accuracy is higher with CUSP-aggregated evidence at every round---64.0\%, 61.0\%, and 63.0\% versus 49.0\%, 60.0\%, and 60.0\%---for gains of 15, 1, and 3 percentage points, respectively.
These results show that supplying the CUSP-selected representative is associated with higher end-to-end system accuracy in this comparison.

\section{Pooling-Weight Ablation}
\label{supp:pooling-ablation}

The uncertainty-weighted variant assigns greater pooling weight to models with lower individual semantic entropy.
When $H_{i,k}>0$ for every model, its pooling rule is

\[
w_{i,k}
=
\frac{H_{i,k}^{-1}}
{\sum_{\ell=1}^{K}H_{i,\ell}^{-1}},
\qquad
\mathbf p_i^{w}=\sum_{k=1}^{K}w_{i,k}\mathbf p_{i,k}.
\]

This direct definition avoids introducing a second symbol $c$; throughout this supplement, $c_{i,\mathcal M}$ is reserved for the hard-answer model-conflict label in Supplementary Section~\ref{supp:conflict-auroc}.
For completeness, let $\mathcal Z_i=\{k:H_{i,k}=0\}$.
The zero-entropy limiting convention is

\[
w_{i,k}=
\begin{cases}
\dfrac{1}{|\mathcal Z_i|}, & k\in\mathcal Z_i \text{ and } |\mathcal Z_i|>0,\\[4pt]
0, & k\notin\mathcal Z_i \text{ and } |\mathcal Z_i|>0,
\end{cases}
\]

while the inverse-entropy definition above applies when $\mathcal Z_i=\varnothing$.
Its normalized uncertainty is

\[
U_i^{w}=
\begin{cases}
\dfrac{H(\mathbf p_i^{w})}{\log_2J_i}, & J_i>1,\\[4pt]
0, & J_i=1.
\end{cases}
\]

The static and multi-agent implementations use $10^{-12}$ and $10^{-8}$, respectively, solely as numerical safeguards that approximate this limiting convention; neither value is part of the formal pooling rule.

\subsection{Static multi-VLM results}
\label{supp:ablation-static-results}

\begin{figure*}[t]
\centering
\begin{minipage}[t]{0.49\textwidth}
\centering
\includegraphics[width=0.85\linewidth]{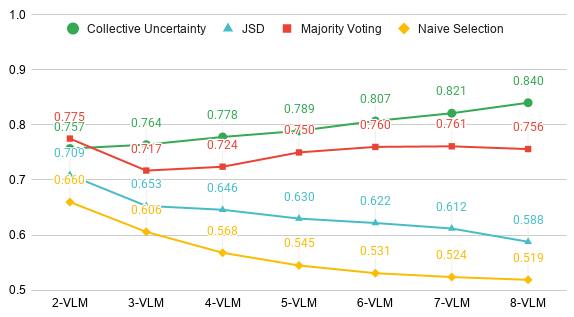}
\par{\footnotesize (a) Prediction-error detection (AUROC)}
\end{minipage}\hfill
\begin{minipage}[t]{0.49\textwidth}
\centering
\includegraphics[width=0.85\linewidth]{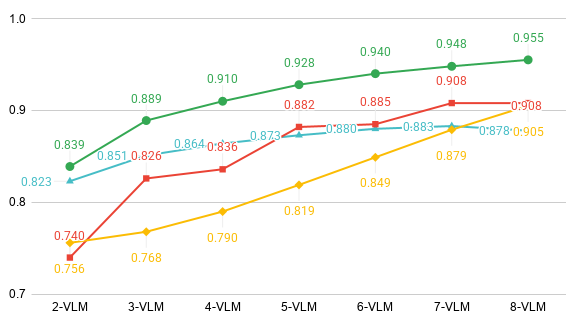}
\par{\footnotesize (b) Abstention ranking (AUARC)}
\end{minipage}
\caption{Small-model scaling for prediction-error detection (AUROC) and abstention ranking (AUARC). Each point macro-averages all systems of size $K$.}
\label{fig:size-scaling}
\end{figure*}

Figure~\ref{fig:size-scaling} provides the ensemble-size context for the small-model results discussed in Section 4.1 of the main paper, while Tables~\ref{tab:ablation-static}--\ref{tab:ablation-latency} compare uncertainty-weighted pooling with CUSP's uniform pooling rule. Table~\ref{tab:ablation-static} matches the principal three-model static UQ setting: uniform pooling performs better in the small-model regime, whereas uncertainty-weighted pooling is slightly stronger in the large and commercial regimes. Table~\ref{tab:ablation-size} extends the comparison across small-model ensemble sizes. Uncertainty weighting has a slight advantage at $K=2$, but uniform pooling leads on both AUROC and AUARC from $K=3$ through $K=8$, with a margin that widens as the ensemble grows. Tables~\ref{tab:ablation-accuracy} and~\ref{tab:ablation-latency} further compare answer accuracy and aggregation overhead. Uniform pooling gives higher macro-averaged accuracy in the small and commercial regimes, while uncertainty-weighted pooling is 0.22 percentage points higher in the large-model regime. Uniform pooling is also faster at every ensemble size, with an across-size mean overhead of 3.97 microseconds compared with 7.45 microseconds for uncertainty-weighted pooling; both remain negligible relative to repeated VLM inference. Taken together, these results support uniform pooling as the default used in the main analysis without implying that it is optimal in every regime.

\begin{table}[t]
\centering
\caption{Static pooling-weight ablation for three-model systems. Values compare uncertainty-weighted collective uncertainty with CUSP's uniform collective uncertainty for prediction-error detection and abstention ranking. \textbf{Bold} marks the best values per row.}
\label{tab:ablation-static}
\footnotesize
\begin{tabular}{llrr}
\toprule
Regime & Metric & Weighted & Uniform \\
\midrule
\multirow{2}{*}{Small}      & AUROC & 0.755 & \textbf{0.764} \\
                            & AUARC & 0.882 & \textbf{0.889} \\
\multirow{2}{*}{Large}      & AUROC & \textbf{0.928} & 0.906 \\
                            & AUARC & \textbf{0.989} & 0.987 \\
\multirow{2}{*}{Commercial} & AUROC & \textbf{0.800} & 0.793 \\
                            & AUARC & \textbf{0.906} & 0.904 \\
\bottomrule
\end{tabular}
\end{table}

\begin{table}[t]
\centering
\caption{Pooling-weight ablation across small-model system sizes. Values report AUROC and AUARC for uncertainty-weighted versus uniform collective uncertainty. \textbf{Bold} marks the best values per row in the same metric.}
\label{tab:ablation-size}
\footnotesize
\begin{tabular}{crrrr}
\toprule
\multirow{2}{*}{$K$-VLM} & \multicolumn{2}{c}{AUROC} & \multicolumn{2}{c}{AUARC} \\
\cmidrule(lr){2-3}\cmidrule(lr){4-5}
 & Weighted & Uniform & Weighted & Uniform \\
\midrule
2 & \textbf{0.783} & 0.757 & \textbf{0.844} & 0.839 \\
3 & 0.755 & \textbf{0.764} & 0.882 & \textbf{0.889} \\
4 & 0.740 & \textbf{0.778} & 0.898 & \textbf{0.910} \\
5 & 0.723 & \textbf{0.789} & 0.907 & \textbf{0.928} \\
6 & 0.715 & \textbf{0.807} & 0.914 & \textbf{0.940} \\
7 & 0.720 & \textbf{0.821} & 0.920 & \textbf{0.948} \\
8 & 0.744 & \textbf{0.840} & 0.926 & \textbf{0.955} \\
\bottomrule
\end{tabular}
\end{table}

\begin{table}[t]
\centering
\caption{Pooling-weight ablation for combination-level answer accuracy. Values are percentages averaged over all non-singleton systems in each regime.}
\label{tab:ablation-accuracy}
\footnotesize
\begin{tabular}{lrr}
\toprule
Regime & Weighted & Uniform \\
\midrule
Small      & 78.01\% & \textbf{79.29\%} \\
Large      & \textbf{94.20\%} & 93.98\% \\
Commercial & 79.68\% & \textbf{79.70\%} \\
\bottomrule
\end{tabular}
\end{table}

\begin{table}[t]
\centering
\caption{Aggregation-overhead ablation for uncertainty-weighted and uniform pooling. Values report additional overhead $\Delta L_{\mu\mathrm{s}}$ above the same VLM UQ sampling latency; lower is faster. \textbf{Bold} marks the best values per row.}
\label{tab:ablation-latency}
\footnotesize
\begin{tabular}{crr}
\toprule
$K$-VLM & Weighted & Uniform \\
\midrule
2 & 4.18 & \textbf{2.56} \\
3 & 5.29 & \textbf{3.06} \\
4 & 6.43 & \textbf{3.56} \\
5 & 7.59 & \textbf{4.03} \\
6 & 8.69 & \textbf{4.49} \\
7 & 9.68 & \textbf{4.95} \\
8 & 10.30 & \textbf{5.11} \\
\midrule
Across-size mean & 7.45 & \textbf{3.97} \\
\bottomrule
\end{tabular}
\end{table}

\subsection{Multi-agent trajectory results}
\label{supp:ablation-trajectory-results}

The trajectory ablation uses the same run and final-failure labels as Section 4.2 of the main paper. Table~\ref{tab:ablation-trajectory} compares uncertainty-weighted and uniform pooling under temporal-mean and temporal-maximum aggregation. Uniform pooling leads in all four comparisons, improving Traj-AUROC and Traj-AUARC by 0.009 and 0.006 under temporal mean and by 0.019 and 0.011 under temporal maximum, respectively. These results support the uniform-pooling choice for the evaluated trajectory while not establishing that uncertainty weighting is universally inferior.

\begin{table}[t]
\centering
\caption{Multi-agent trajectory ablation for uncertainty-weighted and uniform pooling. Values report prediction-error detection (Traj-AUROC) and abstention ranking (Traj-AUARC) under temporal mean and maximum aggregation. \textbf{Bold} marks the best values per row.}
\label{tab:ablation-trajectory}
\footnotesize
\begin{tabular}{llrr}
\toprule
Temporal & Metric & Weighted & Uniform \\
\midrule
\multirow{2}{*}{Mean} & Traj-AUROC & 0.599 & \textbf{0.608} \\
                       & Traj-AUARC & 0.688 & \textbf{0.694} \\
\hdashline
\multirow{2}{*}{Max}  & Traj-AUROC & 0.600 & \textbf{0.619} \\
                       & Traj-AUARC & 0.688 & \textbf{0.699} \\
\bottomrule
\end{tabular}
\end{table}

\section{Implementation Details and Prompts}
\label{supp:implementation}

\subsection{Multi-VLM models and response mapping}
\label{supp:static-models}

\begin{table*}[t]
\centering
\caption{Static multi-VLM model identifiers by regime. Identifiers match the final analysis configurations used for the reported results.}
\label{tab:models}
\footnotesize
\begin{tabular}{lp{0.82\textwidth}}
\toprule
Regime & Models \\
\midrule
Small open-source models &
\path{ministral-3:3b}, \path{qwen3.5:2b}, \path{granite3.2-vision:2b}, \path{qwen3-vl:2b}, \path{llava-phi3:3.8b}, \path{moondream:1.8b}, \path{gemma3:4b}, \path{qwen2.5vl:3b} \\
\addlinespace
Large open-source models &
\path{gemma4:31b}, \path{nemotron3:33b}, \path{qwen3.5:35b}, \path{qwen3.6:35b}, \path{gemma3:27b}, \path{llava:34b} \\
\addlinespace
Commercial models &
\path{anthropic/claude-sonnet-5}, \path{google/gemini-3.1-pro-preview}, \path{minimax/minimax-m3}, \path{moonshot/kimi-k2.6}, \path{openai/gpt-5.6-terra}, \path{xai/grok-4.5} \\
\bottomrule
\end{tabular}
\end{table*}

\textbf{Models.}
The static experiments use the following model identifiers from the final analysis configurations.
We cite the corresponding model-family reports, model cards, or official provider documentation: Ministral-3 \citep{mistral2025mistral3}, Granite Vision \citep{ibm2025granitevision}, Qwen2.5-VL \citep{bai2025qwen25vl}, Qwen3-VL \citep{bai2025qwen3vl}, Qwen3.5 \citep{qwen2026qwen35}, Qwen3.6 \citep{qwen2026qwen36}, Gemma3 and Gemma4 \citep{google2025gemma3,google2026gemma4}, LLaVA-NeXT \citep{liu2024llavanext}, LLaVA-Phi3 \citep{ollama2024llavaphi3}, Moondream2 \citep{korrapati2024moondream2}, Nemotron3 \citep{nvidia2026nemotron3,ollama2026nemotron3}, Claude Sonnet 5 \citep{anthropic2026sonnet5}, Gemini 3.1 Pro Preview \citep{google2026gemini31}, MiniMax M3 \citep{minimax2026m3}, Kimi K2.6 \citep{moonshot2026kimi26}, GPT-5.6 Terra \citep{openai2026gpt56}, and Grok 4.5 \citep{xai2026grok45}.

\textbf{Hardware.}
All locally hosted model inference and experiments were conducted on a DSS 8440 Cauldron server equipped with an Intel Xeon Gold 5218R CPU and eight NVIDIA A40 GPUs with 48 GB of memory each.
All reported latency measurements for locally hosted models were collected on this machine.
Commercial models were accessed through hosted APIs and therefore did not execute on the local hardware.


The open-source endpoints and the commercial APIs that expose decoding control use prediction temperature 0.1 for the main response and sampling temperature 1.0 for the ten UQ samples.
In the evaluated commercial interface, Claude Sonnet 5 \citep{anthropic2026sonnet5} and Kimi K2.6 \citep{moonshot2026kimi26} do not expose temperature control; the client therefore omits it and uses the providers' default decoding behavior.
The other four commercial endpoints use the stated temperature configuration.
This provider-specific limitation reduces direct control over Kimi sampling and should be considered when interpreting the commercial results.
We performed no development-time hyperparameter search. 
The sample count \(N=10\) was fixed a priori based on prior sampling-based semantic uncertainty practice \citep{farquhar2024semantic}.

\subsection{Multi-agent system architecture}
\label{supp:mas-architecture}

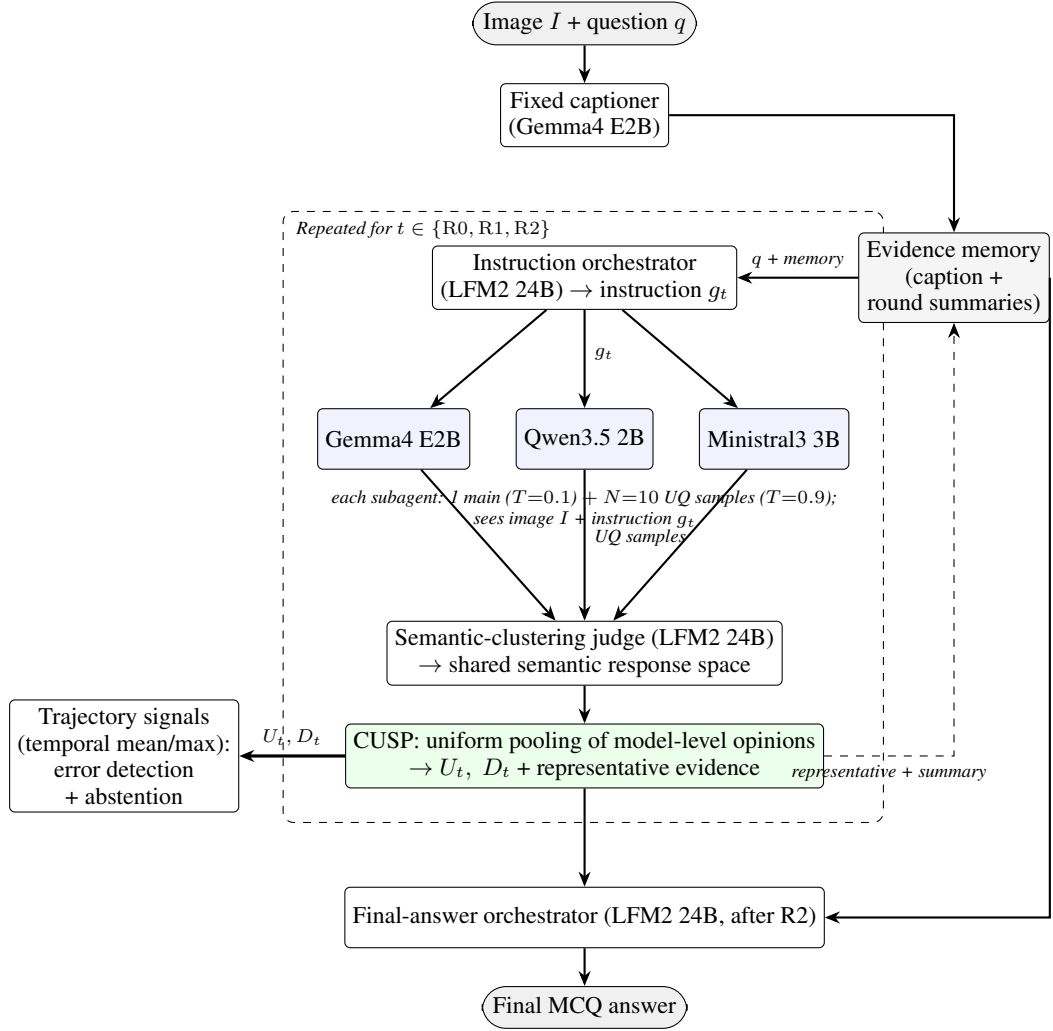
\begin{figure*}[t]
\centering
\begin{tikzpicture}[
  font=\footnotesize,
  >={Stealth[length=2mm,width=1.6mm]},
  proc/.style ={rectangle, rounded corners=2pt, draw, align=center,
                inner sep=3pt, minimum height=8mm, fill=white},
  io/.style   ={rectangle, rounded corners=9pt, draw, align=center,
                inner sep=4pt, fill=black!6},
  agent/.style={proc, fill=blue!5, minimum width=17mm},
  side/.style ={proc, fill=gray!8},
  cusp/.style ={proc, fill=green!8},
  flow/.style ={->, thick},
  feed/.style ={->, dashed},
  sig/.style  ={->, very thick},
  note/.style ={font=\scriptsize\itshape, align=center}
]
\node[io]                        (in)   {Image $I$ + question $q$};
\node[proc, below=5mm of in]     (cap)  {Fixed captioner\\(Gemma4 E2B)};
\node[proc, below=13mm of cap]   (orch) {Instruction orchestrator\\(LFM2 24B) $\to$ instruction $g_t$};
\node[agent, below=13mm of orch] (a2)   {Qwen3.5 2B};
\node[agent, left=6mm of a2]     (a1)   {Gemma4 E2B};
\node[agent, right=6mm of a2]    (a3)   {Ministral3 3B};
\node[note, below=1.5mm of a2]   (an)   {each subagent: 1 main ($T{=}0.1$) $+$ $N{=}10$ UQ samples ($T{=}0.9$);\\ sees image $I$ + instruction $g_t$};
\node[proc, below=11mm of an]    (cl)   {Semantic-clustering judge (LFM2 24B)\\ $\to$ shared semantic response space};
\node[cusp, below=5mm of cl]     (cusp) {CUSP: uniform pooling of model-level opinions\\ $\to$ $U_t,\ D_t$ + representative evidence};
\node[proc, below=13mm of cusp]  (final){Final-answer orchestrator (LFM2 24B, after R2)};
\node[io, below=5mm of final]    (out)  {Final MCQ answer};

\node[side, right=16mm of orch]  (mem)  {Evidence memory\\(caption +\\ round summaries)};
\node[proc, left=14mm of cusp]   (eval) {Trajectory signals\\(temporal mean/max):\\ error detection\\ + abstention};

\draw[flow] (in)  -- (cap);
\draw[flow] (orch) -- (a1);
\draw[flow] (orch) -- node[note,right,pos=0.45]{$g_t$} (a2);
\draw[flow] (orch) -- (a3);
\draw[flow] (a1) -- (cl);
\draw[flow] (a2) -- node[note,right,pos=0.45]{UQ samples} (cl);
\draw[flow] (a3) -- (cl);
\draw[flow] (cl)  -- (cusp);
\draw[flow] (cusp) -- (final);
\draw[flow] (final) -- (out);

\draw[flow] (cap.east) -| (mem.north);
\draw[flow] (mem.west) -- node[note,above]{$q$ + memory} (orch.east);
\draw[feed] (cusp.east) -| node[note,below,pos=0.25]{representative + summary} (mem.south);
\draw[flow] (mem.east) |- (final.east);

\draw[sig] (cusp.west) -- node[note,above]{$U_t, D_t$} (eval.east);

\begin{scope}[on background layer]
  \node[draw, dashed, rounded corners, inner sep=4.5mm,
        fit=(orch)(a1)(a3)(an)(cl)(cusp)] (roundbox) {};
  \node[note, anchor=north west] at (roundbox.north west) {~Repeated for $t\in\{\mathrm{R0,R1,R2}\}$};
\end{scope}
\end{tikzpicture}
\caption{Centralized multi-agent evidence-acquisition architecture for Section 4.2 of the main paper. CUSP is applied to the VLM subagents' UQ samples at each evidence-acquisition round; the final-answer orchestrator is not sampled for uncertainty. Full prompts appear later in this supplement.}
\label{fig:mas-architecture}
\end{figure*}

Figure~\ref{fig:mas-architecture} shows the centralized multi-agent evidence-acquisition system evaluated in Section 4.2 of the main paper.
A fixed captioner supplies scene context to a text-only orchestrator.
Across three evidence-acquisition rounds, the orchestrator issues one focused instruction to three heterogeneous VLM subagents, which independently inspect the image and produce stochastic evidence samples.
The samples are mapped into a shared response space by the semantic-clustering judge and aggregated by CUSP into collective uncertainty, JSD, and a top semantic response class.
The reliability signals are used for trajectory-level evaluation, while the representative evidence is returned to the orchestrator for subsequent rounds and the final answer.

\subsection{Multi-agent models and generation roles}
\label{supp:mas-models}

The three evidence agents are \path{gemma4:e2b-it-q4_K_M}, \path{ministral-3:3b}, and \path{qwen3.5:2b} \citep{google2026gemma4,mistral2025mistral3,qwen2026qwen35}.
The fixed captioner uses the \path{gemma4:e2b-it-q4_K_M} endpoint \citep{google2026gemma4}.
The instruction orchestrator, final-answer orchestrator, and semantic-equivalence judge use \path{lfm2:24b} \citep{liquidai2026lfm224b}.
Each evidence branch produces one main structured response at temperature 0.1 and ten additional UQ samples at temperature 0.9.
The captioner, orchestrator, and semantic judge also use temperature 0.1.
The ten UQ samples, not the main response, define the branch distribution.

\textbf{Free-form clustering and representative selection.}
The free-form implementation pools the structured-evidence samples from all successful VLM responses before clustering.
Two samples are assigned to the same class only if the semantic judge returns true in both comparison directions.
Exact string matches are accepted without a judge call.
If both strings contain explicit integer claims, their extracted integer sets must match before the semantic judge is consulted.
Samples are processed in generation order and compared against members of existing clusters; a sample joins the first cluster containing an equivalent member, otherwise it creates a new cluster.
The first sample that creates a cluster is its representative response.

\textbf{Representative selection and ties.}
In the static experiments, the returned representative is the selected benchmark option.
In the free-form system, the representative is the first response in the selected semantic cluster.
For CUSP, a tie in the uniformly pooled class probabilities is resolved by the earliest cluster index, which follows pooled sample order.
The uncertainty-weighted implementation first selects the lowest-entropy branch and uses that branch's probability among the tied classes; any remaining tie is resolved by the earliest cluster index.
These deterministic rules are reported for reproducibility and are not presented as a methodological contribution.

\textbf{Logarithm bases.}
The static implementation computes entropy and JSD with base-2 logarithms, whereas the multi-agent runtime uses natural logarithms internally.
Because each entropy or divergence is normalized by a logarithm of the same base, the normalized quantities are base-invariant.
For consistency and interpretability, the paper reports unnormalized entropy and divergence in bits.

\begin{table*}[t]
\centering
\caption{Prompt template: semantic-equivalence judgment for response clustering.}
\label{tab:prompt-semantic-clustering}
\begin{promptbox}
\lstset{style=prompttext}
\promptheading{System Prompt}
\begin{lstlisting}
You are a semantic equivalence judge.
\end{lstlisting}
\promptdivider
\promptheading{Prompt}
\begin{lstlisting}[escapeinside={(*@}{@*)}]
Task/context:
(*@\promptplaceholder{context}@*)

Each candidate is structured visual evidence. Decide whether the two candidates express the same visual conclusion for the current evidence request.

Candidate A:
(*@\promptplaceholder{candidate_a}@*)

Candidate B:
(*@\promptplaceholder{candidate_b}@*)

Decision rules:
- Return true only if A and B would belong to the same semantic hypothesis.
- Ignore wording differences and JSON formatting differences.
- Treat matching direct answers with compatible supporting visual evidence as same.
- Return false for conflicting visual facts, counts, objects, attributes, spatial relations, or labels/text.
- Return false if one candidate is definitive and the other says the relevant evidence is not visible or cannot be determined.

Output strictly as JSON:
{"same_semantic_answer": true_or_false, "reason": "short reason"}
\end{lstlisting}
\end{promptbox}
\end{table*}

\begin{table*}[t]
\centering
\caption{Prompt template: extraction of structured visual evidence by each VLM.}
\label{tab:prompt-evidence-extraction}
\begin{promptbox}
\lstset{style=prompttext}
\promptheading{System Prompt}
\begin{lstlisting}
You are a visual evidence extractor.
\end{lstlisting}
\promptdivider
\promptheading{Prompt}
\begin{lstlisting}[escapeinside={(*@}{@*)}]
Answer the evidence instruction using only visible image evidence.
Do not infer the benchmark task or solve a hidden multiple-choice question.
Do not guess hidden causes.
Do not output markdown.
Return ONLY valid JSON with exactly these keys:
{
  "direct_answer": "short answer to the instruction",
  "visual_evidence": ["specific visible fact 1", "specific visible fact 2"],
  "limitations": "what is not visible or ambiguous"
}

EVIDENCE INSTRUCTION:
(*@\promptplaceholder{instruction}@*)
\end{lstlisting}
\end{promptbox}
\end{table*}

\begin{table*}[t]
\centering
\caption{Prompt template: generation of the fixed image caption.}
\label{tab:prompt-image-captioning}
\begin{promptbox}
\lstset{style=prompttext}
\promptheading{System Prompt}
\begin{lstlisting}
You are the fixed image captioner.
\end{lstlisting}
\promptdivider
\promptheading{Prompt}
\begin{lstlisting}
Describe only visible image content.
Mention objects, scene layout, relationships, text, diagrams, or labels if visible.
Do not guess hidden causes.
Do not output markdown.
Return ONLY valid JSON with exactly these keys:
{
  "scene_summary": "one compact paragraph",
  "visible_entities": ["short entity names"],
  "potentially_relevant_details": ["short visible details"]
}
\end{lstlisting}
\end{promptbox}
\medskip
\caption{Prompt template: generation of one focused visual-evidence instruction.}
\label{tab:prompt-evidence-instruction}
\begin{promptbox}
\lstset{style=prompttext}
\promptheading{System Prompt}
\begin{lstlisting}
You are the orchestrator for one visual evidence round.
\end{lstlisting}
\promptdivider
\promptheading{Prompt}
\begin{lstlisting}[escapeinside={(*@}{@*)}]
Ask the VLM evidence extractors for ONE focused piece of visible evidence.
Do not answer the benchmark question here. Do not ask for a final answer letter here.

Rules:
- The first round may be broad but must be relevant to the benchmark question.
- Later rounds must use the prior natural-language round summaries to narrow the next visual check.
- Do NOT repeat or reword any instruction in PREVIOUS INSTRUCTIONS. Each new instruction must ask about a DIFFERENT, more specific visible detail than every previous one.
- If previous rounds were uncertain or disagreeing, ask for the single most decisive visible detail not yet examined.
- Put only exact strings copied from PREVIOUS INSTRUCTIONS into "avoid_repeating". If PREVIOUS INSTRUCTIONS is empty, "avoid_repeating" must be [].

Return ONLY valid JSON with exactly these keys:
{
  "instruction": "one focused, non-repeating visual evidence request",
  "focus": "the specific visible detail this request checks",
  "avoid_repeating": ["exact prior instruction string", "..."]
}

BENCHMARK QUESTION:
(*@\promptplaceholder{benchmark_question}@*)

CAPTION JSON:
(*@\promptplaceholder{caption_json}@*)

PREVIOUS INSTRUCTIONS (do NOT repeat or reword any of these):
(*@\promptplaceholder{previous_instructions}@*)

PREVIOUS ROUND SUMMARY IN NATURAL LANGUAGE:
(*@\promptplaceholder{previous_rounds_summary}@*)
\end{lstlisting}
\end{promptbox}
\end{table*}

\begin{table*}[t]
\centering
\caption{Prompt template: final-answer generation from the caption and pooled VLM evidence.}
\label{tab:prompt-final-answer}
\begin{promptbox}
\lstset{style=prompttext}
\promptheading{System Prompt}
\begin{lstlisting}
You are the final-answer orchestrator.
\end{lstlisting}
\promptdivider
\promptheading{Prompt}
\begin{lstlisting}[escapeinside={(*@}{@*)}]
Make the final decision exactly once using the caption and compact pooled VLM evidence summaries.
Do not request more evidence.
Follow the answer format instruction exactly.
If answer choices are detected, the `answer` value must be one valid option label only. Never use null, None, unknown, a sentence, or visual evidence text as the `answer` value for a multiple-choice question.
Store the explanation separately in `reason`.
Return ONLY valid JSON with exactly these keys:
{
  "answer": "final answer only",
  "reason": "short evidence-grounded reason"
}

ANSWER FORMAT INSTRUCTION:
(*@\promptplaceholder{answer_format_instruction}@*)

BENCHMARK QUESTION:
(*@\promptplaceholder{benchmark_question}@*)

CAPTION JSON:
(*@\promptplaceholder{caption_json}@*)

EVIDENCE INSTRUCTION:
(*@\promptplaceholder{evidence_instruction_json}@*)

POOLED VLM EVIDENCE SUMMARY:
(*@\promptplaceholder{vlm_evidence_summary}@*)

POLICY STOP REASON:
(*@\promptplaceholder{policy_stop_reason}@*)
\end{lstlisting}
\end{promptbox}
\end{table*}

%
%

\end{document}